\documentclass[10pt,twocolumn,letterpaper]{article}
\ifdefined\pdfsuppressptexinfo\pdfsuppressptexinfo=15\relax\fi
\usepackage[pagenumbers]{cvpr}
\usepackage{cuted}
\usepackage{currfile}
\usepackage{caption}

\usepackage{amsthm}
\usepackage{algorithm}
\usepackage{algpseudocode}
\usepackage{multirow}
\usepackage{tabularx}
\usepackage{tikz}
\usetikzlibrary{arrows.meta,positioning}

\newcommand{\missing}{\textemdash{}}

\DeclareRobustCommand{\rev}[1]{#1}
\newcommand{\revcolor}{}

\definecolor{cvprblue}{rgb}{0.21,0.49,0.74}
\usepackage[breaklinks,colorlinks,allcolors=cvprblue]{hyperref}
\title{SELECT: A Submodular Approach for Active LiDAR Semantic Segmentation}
\author{Ruiyu Mao\quad Sarthak Kumar Maharana\quad Xulong Tang\quad Baoming Zhang\quad Yunhui Guo\\
Department of Computer Science, The University of Texas at Dallas}
\hypersetup{pdftitle={SELECT: A Submodular Approach for Active LiDAR Semantic Segmentation},pdfauthor={Ruiyu Mao, Sarthak Kumar Maharana, Xulong Tang, Baoming Zhang, Yunhui Guo}}
\begin{document}
\maketitle
\begin{abstract}
LiDAR semantic segmentation requires costly point annotations, and newly collected scans continually increase this labeling burden. Active learning (AL) reduces this burden by selecting which points to annotate. However, under a limited point-label budget, existing AL methods still struggle to identify representative candidates efficiently, combine model predictions with geometric information from LiDAR scans while reducing redundant queries, and improve performance on classes with few acquired labels. We present SELECT, a new submodular active learning framework that addresses these challenges in two stages.
SELECT first uses LiDAR point features to efficiently select a small, representative candidate pool without clustering.
It then combines local differences in model predictions with LiDAR intensity variation to prioritize queries that complement those already selected. Acquired label counts per class guide this joint selection toward classes with limited supervision. Importantly,  SELECT reuses LiDAR point features and predictions and requires no additional selection network. On SemanticPOSS, SELECT achieves 48.94\% mIoU with SPVCNN using labels for only 0.0127\% of the raw training points, including initialization.
Our candidate screening is also more efficient than the evaluated clustering and Gaussian mixture model baselines.
\end{abstract}

\par
\section{Introduction}
\label{sec:introduction}

LiDAR semantic segmentation assigns a semantic class to each point in a 3D scan, but \rev{obtaining dense human labels is costly}~\cite{xie2023annotator,liu2022less}.
SemanticKITTI, for example, contains over 43,000 scans and about 4.5 billion points, and its annotation required over 1,700 hours of work, including verification and correction~\cite{behley2019semantickitti}.
As new scans are collected, this annotation burden continues to grow, motivating methods that achieve accurate segmentation with fewer human-provided point labels.

\rev{Active learning (AL) reduces annotation costs by iteratively selecting informative locations for labeling using the current model.}
\rev{General-purpose AL methods guide selection using uncertainty~\cite{gal2017deep}, diversity~\cite{sener2018coreset}, or both~\cite{ash2020badge}.} Methods designed specifically for point clouds further exploit local geometry and spatial relationships to guide query selection, as in ReDAL~\cite{wu2021redal}, LiDAL~\cite{hu2022lidal}, and Annotator~\cite{xie2023annotator}. Multi-granularity semi-supervised active learning~\cite{ye2023multigranularity} also incorporates LiDAR intensity information. BaSAL~\cite{wei2023basal} and SSDR-AL~\cite{shao2022ssdr} also account for class imbalance during query selection.

\begin{figure}[!t]
    \centering
    \includegraphics[width=\linewidth]{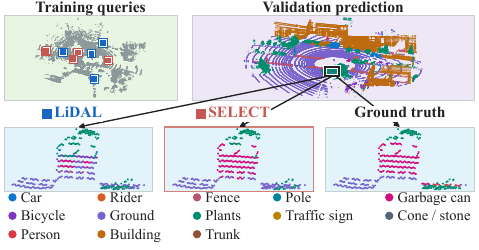}
    \captionof{figure}{\textbf{Sparse annotation and prediction on SemanticPOSS.} Red and blue markers show SELECT and LiDAL queries within the displayed training scan, excluding initialization. The separate validation example was selected retrospectively. SELECT achieves 48.94\% mIoU using 21,278 point labels, while LiDAL achieves 44.33\% using 24,000 point labels.}
    \label{fig:select_teaser}
\end{figure}

Despite these advances, existing AL methods for LiDAR semantic segmentation still face several key challenges. First, candidate selection must scale efficiently to the large number of points in LiDAR scans.
Core-set~\cite{sener2018coreset} and BADGE~\cite{ash2020badge} use sample representations to select diverse samples, while ReDAL~\cite{wu2021redal} clusters region features to reduce redundant queries. These operations can become computationally expensive over large candidate pools. Second, query selection should combine informative cues while accounting for batch redundancy. However, existing methods emphasize different aspects of this goal. LiDAL~\cite{hu2022lidal} reduces redundancy by removing corresponding regions across frames, while multi-granularity semi-supervised active learning~\cite{ye2023multigranularity} combines prediction entropy with LiDAR intensity to identify informative queries. Combining cues alone does not ensure a complementary batch, since high-scoring points may provide overlapping information. This motivates a joint objective that evaluates prediction and intensity cues through each candidate’s additional contribution to the selected batch. Third, query selection should account for how many points have already been labeled for each class. However, BaSAL~\cite{wei2023basal} allocates budgets across size-based partitions, while SSDR-AL~\cite{shao2022ssdr} weights queries using the dominant-class frequencies of superpoints. These criteria may not reflect the distribution of acquired point labels.

To address these three challenges, we propose SELECT, a new \underline{S}ubmodular activ\underline{e} learning framework for \underline{L}iDAR s\underline{e}manti\underline{c} segmen\underline{t}ation.
SELECT first constructs a small representative candidate pool for each LiDAR scan.
Its submodular coverage objective favors candidates whose features are not yet well covered, while stochastic sampling limits the number evaluated at each step.
This provides efficient screening without feature clustering and reduces the cost of subsequent joint selection. Within this smaller pool, SELECT integrates intensity-weighted feature coverage, predicted-class diversity, and class support into a joint submodular objective.
Local prediction differences can signal confusion, while our SemanticPOSS analysis links greater intensity variation to more frequent local depth transitions (Sec.~\ref{sec:ablation}). Cumulative true class-label counts further indicate the supervision available for each class, guiding selection toward classes with fewer labels.
\rev{Intensity-weighted coverage and class support account for what the batch already covers, while predicted-class diversity contributes a fixed local score.}
Each query’s additional contribution is then divided by the number of new point labels it requires, accounting for annotation cost under the limited budget. SELECT reuses features and predictions from the segmentation backbone without requiring an additional selection network, making it compatible with different backbones that provide these outputs.

Our contributions are as follows.
\begin{itemize}
    \item \textbf{Efficient representative screening.}
    We use stochastic submodular selection to form a compact representative candidate pool without feature clustering, reducing the work required for joint selection.

    \item \textbf{Complementary query selection.}
    Our joint objective combines local prediction differences with intensity-weighted feature coverage to favor informative queries that complement the selected batch.

    \item \textbf{Class support from acquired labels.}
    Within the same objective, we use the distribution of acquired labels to favor classes with few labels.
\end{itemize}

\section{Related Work}
\label{sec:related_work}

\subsection{Label-Efficient 3D Semantic Segmentation}

Sparse and semi-supervised learning reduce the labels needed to train point-cloud segmentation models.
SQN learns from sparse point annotations by gathering neighboring features from representations of the full point cloud~\cite{hu2022sqn}.
LESS requests one point label for each class present in a pre-segmented component and derives weak and propagated labels for training~\cite{liu2022less}.
LaserMix mixes laser-beam regions from labeled and unlabeled scans and uses pseudo-labels and prediction consistency for training~\cite{kong2023lasermix}.
These studies show that dense predictions can be learned from sparse human supervision.
SELECT focuses on choosing where to acquire that supervision, while using surrounding unlabeled points as spatial context.

\subsection{Active Learning for Point Clouds}
Since point-cloud segmentation models can learn from sparse labels, active learning can reduce annotation costs by selecting informative locations for human labeling. Among general active learning methods, Core-set covers the feature space, while BADGE combines uncertainty and diversity via gradient embeddings~\cite{sener2018coreset,ash2020badge}.
NEAT uses prediction--feature inconsistency for open-set annotation~\cite{mao2024neat}.
For point clouds, ReDAL clusters region features to reduce redundant queries, while LiDAL selects regions using prediction differences across aligned scans~\cite{wu2021redal,hu2022lidal}.
\rev{HPAL selects individual points using uncertainty across spatial scales, whereas Annotator uses voxel confusion degree (VCD) to measure predicted-class diversity~\cite{xu2023hpal,xie2023annotator}.}
Multi-granularity semisupervised active learning combines prediction entropy with LiDAR intensity~\cite{ye2023multigranularity}. BaSAL allocates budgets across size-based partitions, while SSDR-AL weights queries using the dominant-class frequencies of superpoints~\cite{wei2023basal,shao2022ssdr}.
Building on these advances, SELECT pairs efficient representative screening with joint optimization of intensity-weighted coverage, prediction differences, and label balance.

\subsection{Submodular Data Selection}

Submodular coverage, often used to select representative data, measures what a candidate adds to the selected set and gives smaller gains for repeated coverage~\cite{wei2015submodularity}.
Wei et al.~\cite{wei2015submodularity} propose FASS, which first filters candidates by uncertainty and then selects a batch using a submodular objective.
Mirzasoleiman et al.~\cite{mirzasoleiman2015lazier} reduce objective evaluations in monotone submodular maximization under a subset-size limit by sampling candidates at each step.
For active 3D object detection, STONE uses submodular selection to address coverage and imbalance, while LH3D considers learnability in roadside scenes~\cite{mao2024stone,mao2026lh3d}.
SELECT adapts these tools to acquisition under a point-label budget.
It uses stochastic selection with feature coverage to screen candidates, then combines intensity-weighted coverage, predicted-class diversity, and class support in a joint objective.
We compare marginal gains per new point label.

\section{Background}
\label{sec:background}

\paragraph{LiDAR semantic segmentation.}
LiDAR semantic segmentation assigns a semantic class to each point in a point cloud.
Let $\mathcal{P}_i=\{\mathbf{p}_{ij}\}_{j=1}^{n_i}$ denote the $i$-th scan, where $\mathbf{p}_{ij}=(x_{ij},y_{ij},z_{ij},r_{ij})$ represents a point with 3D coordinates and measured return intensity.
Each point has a target label $c_{ij}\in\{0,\ldots,C\}$. The labels $1,\ldots,C$ denote the semantic classes, and $0$ denotes the ignored label.
A segmentation model $f_\theta$ processes the scan to produce a feature vector $\mathbf{e}_{ij}$ and a predicted class $\hat c_{ij}$ for each point.
With sparse annotations, the model uses surrounding unlabeled points as context and receives supervision only from labeled points with valid semantic labels.

\paragraph{Active learning with a point-label budget.}
Under a limited point-label budget, active learning must consider both where to annotate and how many new labels each query requires.
We identify each raw point by its scan and point indices $(i,j)$.
Let $L_t$ and $U_t$ denote the labeled and unlabeled raw-point indices before query round $t$.
Starting from a small labeled set $L_0$, the model is trained and used to guide subsequent queries.

Voxels group nearby points into local cells and are widely used by LiDAR segmentation backbones~\cite{choy20194d,tang2020searching}.
\rev{Following Annotator~\cite{xie2023annotator}, we use $0.25$\,m voxels as local annotation candidates.}
For voxel $v$, $\mathcal{T}_v$ contains the raw-point indices retained for that candidate.
At AL round $t$, $\mathcal{U}_{t,v}=\mathcal{T}_v\cap U_t$ contains the indices of these points that are still unlabeled.
Labeling $v$ costs $\kappa_v=|\mathcal{U}_{t,v}|$ new point labels.

In round $t$, we label up to $b_t$ new points, where $b_t\leq B-|L_t|$.
Let $Q_t\subseteq U_t$ be these points.
We add them to the labeled set:
\begin{equation}
    L_{t+1}=L_t\cup Q_t,
    \qquad |L_{t+1}|\leq B.
    \label{eq:background_label_budget}
\end{equation}
The budget $B$ counts all acquired point labels, including initialization and ignored labels.
Before round $t$, $\mathbf{h}_t$ records how many true labels have been acquired for each valid class, including initialization.
After annotation, we update these counts to $\mathbf{h}_{t+1}$ and train the model with the expanded labeled set.

\paragraph{Submodular selection.}
A query can be informative on its own yet repeat information already covered by the selected batch.
Submodular coverage accounts for this redundancy through diminishing returns~\cite{wei2015submodularity}.
Let $\mathcal{V}$ be a finite candidate set and $F:2^{\mathcal{V}}\rightarrow\mathbb{R}$ assign a value to each subset.
The marginal gain of adding $v$ to $S$ is
\begin{equation}
    \Delta_F(v\mid S)=F(S\cup\{v\})-F(S).
    \label{eq:background_marginal_gain}
\end{equation}
The function is submodular if
\begin{equation}
    \Delta_F(v\mid A)\geq\Delta_F(v\mid T),
    \label{eq:background_submodularity}
\end{equation}
for all $A\subseteq T\subseteq\mathcal{V}$ and $v\in\mathcal{V}\setminus T$.
It is monotone if $F(A)\leq F(T)$ whenever $A\subseteq T$.

We use a standard submodular coverage function $\Psi$ that gives smaller gains for repeated coverage~\cite{wei2015submodularity,iyer2019memoization}:
\begin{equation}
    \Psi(S;\mathbf{a})=
    \sum_{j=1}^{m}
    \log\left(1+\sum_{v\in S}a_{v,j}\right).
    \label{eq:background_log_coverage}
\end{equation}
Here, $\mathbf{a}=\{\mathbf{a}_v\}_{v\in\mathcal{V}}$ is fixed during selection.
Each entry $a_{v,j}\geq0$ is candidate $v$'s contribution to attribute $j$, such as a feature dimension or a class.
The inner sum measures how much of that attribute is covered by $S$.
The logarithm gives a larger gain for the same positive contribution when the attribute is less covered.
Thus, $\Psi$ is normalized, monotone, and submodular.
Caching these sums avoids summing over $S$ again when computing marginal gains~\cite{iyer2019memoization}.

For fixed scores $s_v\geq0$, the additive function $M(S)=\sum_{v\in S}s_v$ is modular: adding $v$ always gives a gain of $s_v$.
Modular functions are also submodular.
Fixed nonnegative combinations of $M$ and $\Psi$ remain monotone and submodular.
In SELECT, $\Psi$ measures feature coverage for candidate screening, intensity-weighted coverage for joint selection, and support for classes with few acquired labels.
The modular term $M$ measures local predicted-class diversity.
The next section defines these inputs and the query-selection rules.

\section{Method}
\label{sec:method}

\begin{figure*}[t]
    \centering
    \includegraphics[width=0.9\textwidth]{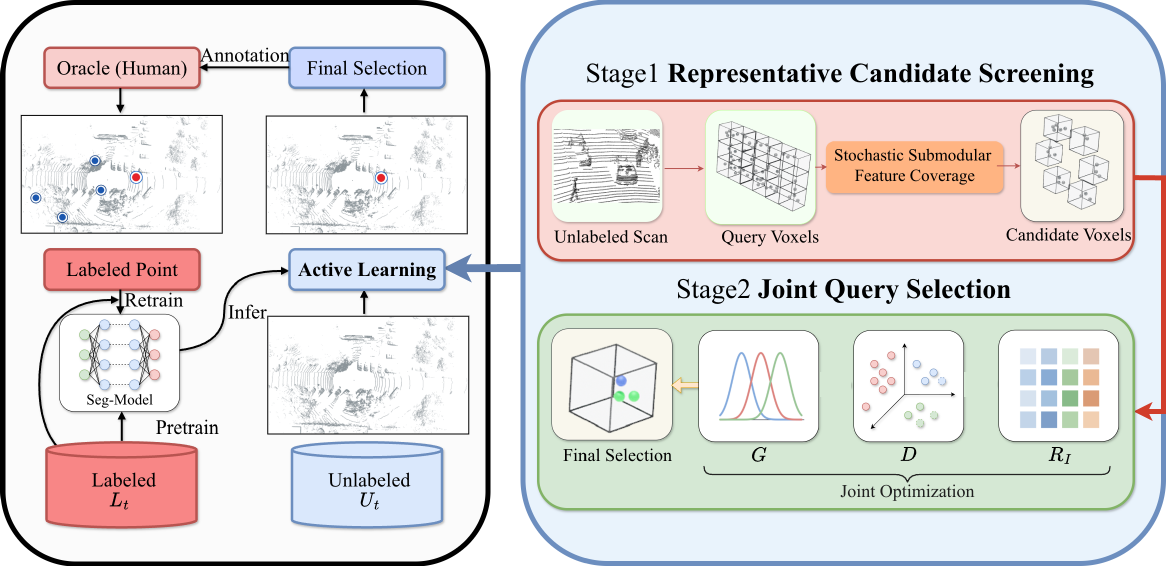}
    \caption{\textbf{Overview of SELECT.} Stage~1 screens representative candidate voxels using model features and stochastic submodular selection. Stage~2 combines intensity-weighted feature coverage, local prediction diversity, and class support from acquired labels, then selects queries by marginal gain per new point label under point-budget and per-scan constraints. Newly annotated points are used to update the labeled set and train the segmentation model.}
    \label{fig:pipeline}
\end{figure*}

Small query units provide precise annotation control but create large candidate pools.
Jointly evaluating feature coverage, local prediction differences, and class support over the full pool requires repeated gain evaluations.
SELECT therefore screens representative candidates before joint selection (Fig.~\ref{fig:pipeline}).
The joint objective combines intensity-weighted feature coverage, predicted-class diversity, and acquired class-label counts to favor complementary queries and support classes with few labels.
Queries are ranked by marginal gain per new point label.
SELECT reuses backbone features and predictions without an additional selection network.
Algorithm~\ref{alg:select} in the supplementary material details one acquisition round.

\subsection{Stage 1: Representative Candidate Screening}
\label{sec:method_screening}

Screening must preserve feature coverage while limiting computation.
ReDAL reduces redundancy through feature clustering~\cite{wu2021redal}; we instead use submodular coverage to favor features less covered by the retained set.
In our ablation, the tested $k$-means variants took over $100\times$ as long to screen candidates as submodular screening, with comparable accuracy. Random screening was faster but achieved lower mIoU (Table~\ref{tab:resources}).

For scan $\mathcal{P}_i$, let $\mathcal{V}_{t,i}$ contain candidates with $\mathcal{U}_{t,v}\neq\varnothing$.
Let $\mathbf{e}_{ij}\in\mathbb{R}^{d_f}$ be the point feature from $f_\theta$ immediately before its final classifier.
We average these features over the retained members $\mathcal{T}_v$, apply a fixed Gaussian projection $\mathbf{M}\in\mathbb{R}^{d_f\times d_p}$, and form a nonnegative coverage vector:
\begin{equation}
\begin{aligned}
    \mathbf{z}_v
        &= \frac{1}{|\mathcal{T}_v|}
        \sum_{(i,j)\in\mathcal{T}_v}\mathbf{e}_{ij},\\
    \mathbf{g}_v
        &= [(\mathbf{z}_v\mathbf{M})_+,
            (-\mathbf{z}_v\mathbf{M})_+],\\
    \boldsymbol{\phi}_v
        &= \frac{\mathbf{g}_v}
        {\max(\|\mathbf{g}_v\|_1,\eta)}
        \in\mathbb{R}_+^d .
\end{aligned}
\label{eq:method_candidate_features}
\end{equation}
Here, $(\cdot)_+$ is the elementwise positive part, $d=2d_p$, and $\eta>0$ prevents division by zero.
Feature averaging includes any retained labeled points; intensity statistics, predicted class counts, and annotation costs use only $\mathcal{U}_{t,v}$.
Projection and retention settings are given in supplementary Sec.~\ref{sec:select_implementation}.

Using $\Psi$ from Eq.~\eqref{eq:background_log_coverage}, we screen $A\subseteq\mathcal{V}_{t,i}$ by
\begin{equation}
    a^{0}_{v,k}=\frac{\phi_{vk}}{\tau_f},
    \qquad
    R_0(A)=\Psi(A;\mathbf{a}^{0}),
    \label{eq:method_screening_objective}
\end{equation}
where $\tau_f>0$ controls the feature scale.
\rev{We compute marginal gains from cached feature sums, without constructing a pairwise similarity matrix~\cite{iyer2019memoization}.}
At each step, stochastic greedy~\cite{mirzasoleiman2015lazier} samples a subset of remaining candidates, with sample size controlled by $\varepsilon$, and retains the sampled candidate with the largest gain in $R_0$.
For each scan, we keep all candidates if there are at most $K$; otherwise, we select $K$ candidates.
The selected candidates from all scans form the Stage~2 pool $\mathcal{C}_t$.

\subsection{Stage 2: Joint Query Selection}
\label{sec:method_joint}

Feature coverage alone does not explicitly account for local prediction differences or acquired class-label counts.
We combine intensity-weighted feature coverage ($R_I$), local predicted class diversity ($D$), and a term for balancing supervision across classes ($G$):
\begin{equation}
    F_t(S)=
    \frac{R_I(S)}{Z_R}
    +\frac{D(S)}{Z_D}
    +\frac{G(S;\mathbf{h}_t)}{Z_G},
    \label{eq:method_joint_objective}
\end{equation}
for $S\subseteq\mathcal{C}_t$.
The positive scales $Z_R,Z_D,Z_G$ and parameter settings are defined in supplementary Sec.~\ref{sec:select_method_details}.
Features, predictions, intensity weights, class history, and scales remain fixed during batch selection.

\begin{figure}[t]
    \centering
    \includegraphics[width=0.9\columnwidth,trim=5pt 2pt 5pt 1pt,clip]{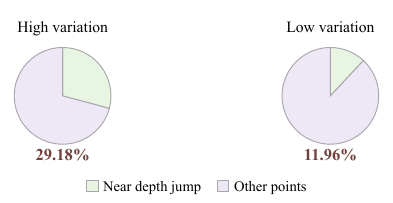}
    
    \caption{Voxels with high intensity variation have a higher fraction of points near detected depth jumps.}
    \label{fig:stage2_intensity_depth}
\end{figure}

\paragraph{Intensity-weighted feature coverage.}
Our SemanticPOSS diagnostic associates intensity variation with local depth structure (Fig.~\ref{fig:stage2_intensity_depth}).
\rev{Among candidates with multiple retained new points, the mean fraction of points near the depth-jump proxy is $29.18\%$ for high intensity variation and $11.96\%$ for low variation. The low-variation group is reweighted by new-point count, range, and dominant ground-truth class.}
This observation motivates weighting feature coverage by intensity variation.

Let $\delta_v$ be twice the population standard deviation of raw intensities in $\mathcal{U}_{t,v}$ after scaling to $[0,1]$.
We define
\begin{equation}
\begin{aligned}
    \omega_v &=
    \frac{1+\delta_v}{1+\max_{u\in\mathcal{C}_t}\delta_u},\\
    a^{I}_{v,k}&=\frac{\omega_v\phi_{vk}}{\tau_f},
    \qquad R_I(S)=\Psi(S;\mathbf{a}^{I}).
\end{aligned}
    \label{eq:method_weighted_coverage}
\end{equation}
Intensity variation increases the feature weight, while repeated coverage reduces marginal gains.

\begin{figure}[t]
    \centering
    \includegraphics[width=0.9\columnwidth,trim=5pt 9pt 5pt 9pt,clip]{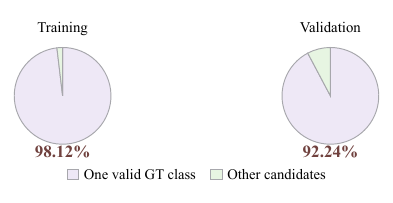}
    
    \caption{\rev{Prediction disagreement can occur in candidates whose retained points share one ground-truth class.}}
    \label{fig:stage2_prediction_disagreement}
\end{figure}

\paragraph{Local predicted class diversity.}
\rev{On 60 SemanticPOSS training scans, $98.12\%$ of candidates with multiple retained new points contain only one valid ground-truth class, excluding ignored labels. In a separate validation diagnostic, $92.24\%$ of 5,970 high-disagreement candidates have all retained new points in the same valid ground-truth class (Fig.~\ref{fig:stage2_prediction_disagreement}).}
\rev{Prediction disagreement can therefore occur within a single ground-truth class.}

Following Annotator~\cite{xie2023annotator}, we use the entropy of predicted class proportions.
Let $n_{v,c}$ count points in $\mathcal{U}_{t,v}$ predicted as valid class $c$, and let $n_v=\sum_{c=1}^{C}n_{v,c}$.
For $n_v>0$, set $q_{v,c}=n_{v,c}/n_v$ and define
\begin{equation}
\begin{aligned}
    D_v &=
    -\frac{n_v}{\log C}
    \sum_{c=1}^{C}q_{v,c}\log q_{v,c},\\
    D(S)&=\sum_{v\in S}D_v.
\end{aligned}
    \label{eq:method_prediction_diversity}
\end{equation}
\rev{This score weights normalized entropy by valid-prediction count, with $D_v=0$ for $n_v=0$ and $0\log0=0$. Fixed scores make $D$ modular; each vanishes when all valid predictions agree, including for a single new point.}

\begin{figure}[t]
    \centering
    \includegraphics[width=\columnwidth]{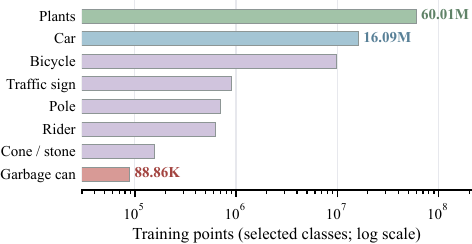}
    
    \caption{\textbf{SemanticPOSS training class distribution.} Raw-point counts for eight selected valid classes in 2,488 training scans. Plants contain about 675 times as many points as garbage cans.}
    \label{fig:stage2_class_distribution}
\end{figure}

\paragraph{Balancing supervision across classes.}
\rev{The SemanticPOSS training set contains about 675 times as many points labeled as \textit{plants} as points labeled as \textit{garbage cans}} (Fig.~\ref{fig:stage2_class_distribution}).
We therefore track the true labels acquired for each class.

\begin{figure}[t]
    \centering
    \includegraphics[width=0.75\columnwidth]{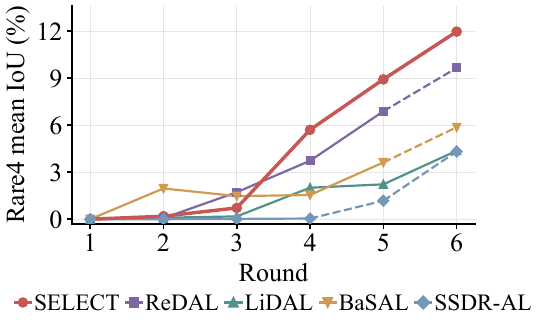}
    
    \caption{Rare4 IoU (rider, pole, garbage can, and cone/stone) across SemanticPOSS query rounds. \rev{Validation epochs are 2, 6, 8, 12, 14, and 18; dashed segments indicate no new labels. Actual label counts differ across methods.}}
    \label{fig:rare4_query_rounds}
\end{figure}

BaSAL allocates budgets across size-based partitions~\cite{wei2023basal}; we scale predicted class counts by acquired true-label counts:
\begin{equation}
\begin{aligned}
    a^{B}_{v,c}&=\frac{n_{v,c}}{\tau_b+h_{t,c}},\\
    G(S;\mathbf{h}_t)&=\Psi(S;\mathbf{a}^{B}),
\end{aligned}
    \label{eq:method_class_support}
\end{equation}
Here, $h_{t,c}$ counts acquired true point labels for class $c$, including initialization, and $\tau_b>0$ provides a positive baseline.
For the same predicted count, a smaller $h_{t,c}$ gives a larger contribution.
Marginal gains decrease as the batch accumulates predicted points of that class.
In our ablation, removing $G$ lowers Rare4 IoU; Fig.~\ref{fig:rare4_query_rounds} compares methods across query rounds.

\begin{table*}[t]
\caption{\textbf{SemanticPOSS validation IoU (\%) at epoch~60.}
SELECT used 21,278 labels.
\rev{All methods use a 24k-point budget.}
Bold marks the best SPVCNN score per column; \rev{SQN uses a separate architecture.}}
\label{tab:budget_results}
\label{tab:semanticposs_spvcnn_24k_classes}
\centering
\begingroup
\fontsize{8}{9.3}\selectfont
\setlength{\tabcolsep}{2.6pt}
\setlength{\arrayrulewidth}{0.25pt}
\setlength{\doublerulesep}{1pt}
\renewcommand{\arraystretch}{1.17}
\def\SelectPossHead#1#2{\shortstack{\rule[-2pt]{0pt}{10pt}\textbf{#1}\\\strut\textbf{#2}}}
\resizebox{\linewidth}{!}{%
\begin{tabular}{l|*{13}{c|}c}
\toprule
\textbf{Method} & \textbf{Car} & \textbf{Bike} & \textbf{Person} & \textbf{Rider} &
\textbf{Ground} & \textbf{Bldg.} & \textbf{Fence} & \textbf{Plants} & \textbf{Trunk} &
\textbf{Pole} & \SelectPossHead{Traffic}{sign} &
\SelectPossHead{Garbage}{can} & \SelectPossHead{Cone/}{stone} &
\textbf{mIoU} \\ \hline
Random & 50.41 & 49.93 & 38.55 & 10.94 & \textbf{81.75} & \textbf{71.97} & 49.45 & 77.37 & 23.99 & 29.48 & 12.96 & 0.50 & 0.00 & 38.25 \\ 
Entropy & 34.45 & 44.47 & 36.75 & 27.50 & 79.99 & 63.68 & 44.63 & 75.94 & 6.84 & 23.24 & 29.89 & 0.00 & 0.00 & 35.95 \\ 
\rev{BGADL}~\cite{tran2019bgadl} & 29.13 & 37.87 & 44.42 & 9.98 & 71.01 & 55.21 & 45.11 & 74.46 & 25.84 & 7.31 & 5.56 & 4.13 & 6.67 & 32.05 \\ 
BADGE~\cite{ash2020badge} & 42.26 & 44.40 & 48.70 & 47.19 & 81.10 & 69.39 & 50.14 & 73.85 & 31.54 & 27.37 & 26.45 & 0.00 & 0.00 & 41.72 \\ 
Core-Set~\cite{sener2018coreset} & 40.38 & 43.00 & 40.99 & 34.23 & 77.92 & 70.68 & 43.13 & 72.81 & 26.04 & 27.43 & 17.98 & 0.01 & 1.49 & 38.16 \\ 
ReDAL~\cite{wu2021redal} & 36.97 & 38.53 & 53.09 & 52.53 & 79.31 & 66.70 & 54.56 & 74.22 & 15.19 & 24.50 & 29.67 & \textbf{11.45} & 27.17 & 43.38 \\ 
SQN~\cite{hu2022sqn} & 56.50 & 51.69 & 50.22 & 44.97 & 82.97 & 71.22 & 51.94 & 78.19 & 35.98 & 23.60 & 22.53 & 0.00 & 0.00 & 43.83 \\ 
BaSAL~\cite{wei2023basal} & 30.31 & 49.18 & 51.07 & 45.48 & 76.89 & 67.75 & 47.96 & 76.40 & 27.08 & 29.42 & 30.72 & 7.10 & 7.41 & 42.06 \\ 
LiDAL~\cite{hu2022lidal} & 46.96 & 43.24 & 48.38 & 54.24 & 81.19 & 68.77 & \textbf{57.04} & 75.42 & \textbf{33.72} & 30.65 & 14.28 & 3.18 & 19.27 & 44.33 \\
MILAN-derived~\cite{samet2024milan} & 35.50 & 41.08 & 25.75 & 16.13 & 78.00 & 63.14 & 25.66 & 70.11 & 12.94 & 22.37 & 10.76 & 0.00 & 0.00 & 30.88 \\ 
SSDR-AL~\cite{shao2022ssdr} & 32.67 & 42.06 & 24.17 & 21.30 & 76.82 & 63.25 & 37.65 & 74.97 & 22.37 & 20.13 & 17.20 & 0.00 & 0.00 & 33.28 \\ 
Margin~\cite{joshi2009multiclass} & 43.14 & \textbf{52.58} & 44.73 & 48.47 & 81.28 & 70.69 & 52.73 & 77.91 & 19.33 & 32.63 & 11.19 & 0.00 & 0.63 & 41.18 \\ 
VCD (Annotator)~\cite{xie2023annotator} & 31.96 & 44.47 & 36.29 & 27.05 & 79.71 & 63.15 & 46.79 & 75.01 & 8.78 & 26.69 & 32.81 & 4.83 & 0.00 & 36.73 \\ 
HPAL~\cite{xu2023hpal} & 26.62 & 52.22 & 34.66 & 26.85 & 80.68 & 71.88 & 53.79 & \textbf{78.57} & 0.76 & 23.19 & 1.46 & 0.00 & 0.00 & 34.67 \\ 
LaserMix~\cite{kong2023lasermix} & 34.13 & 4.00 & 0.11 & 0.20 & 79.56 & 68.42 & 0.15 & 68.29 & 0.01 & 0.07 & 13.97 & 0.03 & 0.01 & 20.69 \\ 
LESS~\cite{liu2022less} & \textbf{58.11} & 50.99 & \textbf{55.64} & 48.34 & 76.90 & 70.32 & 52.59 & 75.37 & 8.82 & 26.97 & 27.74 & 4.22 & 22.43 & 44.50 \\ \hline
\textbf{SELECT} & 40.33 & 50.97 & 54.45 & \textbf{54.34} & 80.78 & 70.00 & 55.89 & 78.40 & 27.46 & \textbf{34.82} & \textbf{35.68} & 9.46 & \textbf{43.70} & \textbf{48.94} \\ \bottomrule
\end{tabular}%
}
\endgroup

\end{table*}

\begin{figure*}[t]
    \centering
    \includegraphics[width=\textwidth]{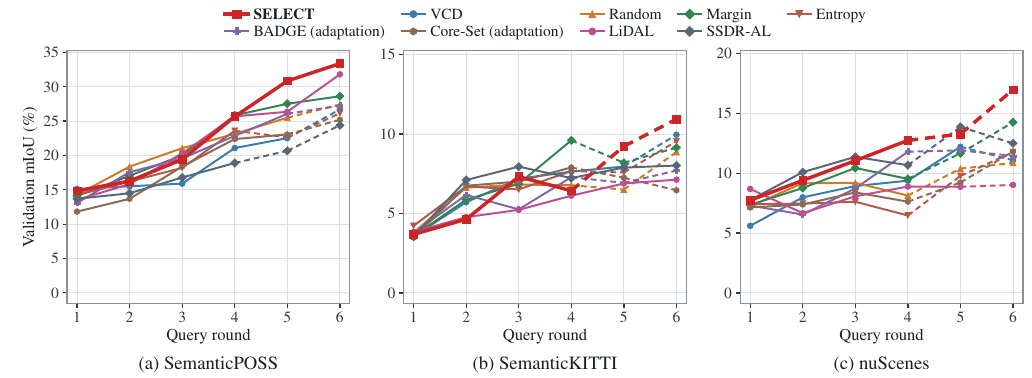}
    \caption{\textbf{Validation mIoU across query rounds with SPVCNN.} Round 1 is initialization; rounds 2--6 follow acquisition at epochs 4, 7, 10, 13, and 16. Validation uses epochs 2, 6, 8, 12, 14, and 18. Dashed segments indicate rounds without new annotations. \rev{These curves stop at epoch~18; comparison tables use epoch~60.}}
    \label{fig:query_round_miou_spvcnn}
\end{figure*}

\paragraph{Query selection.}
With annotation cost $\kappa_v=|\mathcal{U}_{t,v}|$, we score candidates by joint marginal gain per new point label:
\begin{equation}
    \rho_t(v\mid S)=
    \frac{F_t(S\cup\{v\})-F_t(S)}{\kappa_v}.
    \label{eq:method_gain_per_cost}
\end{equation}
Starting from $S=\varnothing$, we repeatedly add the highest-scoring feasible candidate using lazy marginal-gain updates.
A candidate is feasible if its full cost fits the remaining round allowance and its scan has no selected candidate.
Selection stops when no feasible candidate remains, possibly leaving unused budget.
Tie-breaking details are given in supplementary Sec.~\ref{sec:select_method_details}.
We request labels for $Q_t=\bigcup_{v\in S}\mathcal{U}_{t,v}$ and update the labeled set and class history.

\section{Experiments}
\label{sec:experiments}
\subsection{Experimental Setup}

\paragraph{\rev{Datasets.}}
\rev{We evaluate on SemanticPOSS~\cite{pan2020semanticposs}, nuScenes~\cite{fong2022panoptic}, and SemanticKITTI~\cite{behley2019semantickitti}.}

\paragraph{\rev{Backbones and evaluation.}}
\rev{We evaluate SELECT with SPVCNN~\cite{tang2020searching} and MinkNet~\cite{choy20194d}, using coordinates and LiDAR intensity as input. Comparison tables report single-run mIoU at epoch~60. \rev{We evaluate raw-point IoU over 13, 16, and 19 valid classes on SemanticPOSS, nuScenes, and SemanticKITTI, respectively. Ignored ground-truth points are excluded; ignore predictions on valid points count as false negatives.} On SemanticPOSS, Rare4 is the mean IoU over rider, pole, garbage can, and cone/stone. Further details are in supplementary Sec.~\ref{sec:supp_experiment_setup}.}

\paragraph{\rev{Annotation budgets.}}
\rev{SELECT and the four shared voxel baselines---random, entropy, margin, and VCD---start with the same 6,000 labeled points per dataset. Budgets are 12k, 18k, and 24k on SemanticPOSS and 24k on nuScenes and SemanticKITTI, including initialization and acquired ignore labels. Actual counts may fall below the budget and differ across methods. The acquisition schedule, budget protocol, and per-method label counts are in supplementary Sec.~\ref{sec:supp_experiment_setup}.}

\paragraph{\rev{Baseline adaptations.}}
\rev{Baseline selection and training protocols are adapted to these point-label budgets (supplementary Sec.~\ref{sec:supp_baselines}). We compare SQN~\cite{hu2022sqn}, LaserMix~\cite{kong2023lasermix}, and LESS~\cite{liu2022less}. SQN uses a sparse Minkowski encoder with a four-scale query head and 24,000 fixed labels from epoch~1. Sparse-point LaserMix adds teacher-generated labels; LESS also requests class information for pre-segmented components.}

\subsection{Results}
\label{sec:budget_comparison}
\rev{Figure~\ref{fig:query_round_miou_spvcnn} reports SPVCNN validation mIoU at epochs 2, 6, 8, 12, 14, and 18. SELECT leads on all three datasets at the last plotted checkpoint; comparison tables report epoch~60 results.}

\paragraph{\rev{Results on SemanticPOSS.}}
\rev{Table~\ref{tab:budget_results} reports class-wise IoU at epoch~60 under a 24k-point budget. SELECT achieves 48.94\% mIoU with 21,278 labels and the highest Rare4 mean IoU. It scores lower than LESS on car. Margin achieves 52.58\% IoU on bicycle, while Random and SQN lead on building and ground, respectively.}

\paragraph{\rev{Results across budgets.}}
\rev{Table~\ref{tab:budget_comparison} compares single-seed SPVCNN results at epoch~60 under 12k, 18k, and 24k budgets. SELECT achieves the highest mIoU at each budget; actual label counts and training protocols differ.}

\begin{table}[t]
\caption{\rev{\textbf{SemanticPOSS mIoU (\%) across budgets.} Single-seed SPVCNN results at epoch~60. Budgets include initialization; actual label counts and training protocols differ.}}
\label{tab:budget_comparison}
\centering
\revcolor
\small
\begin{tabular}{lrrr}
\toprule
Method & 12k & 18k & 24k \\
\midrule
Random & 35.16 & 36.71 & 38.25 \\
ReDAL & 40.55 & 42.08 & 43.38 \\
LiDAL & 37.54 & 39.69 & 44.33 \\
BaSAL & 35.71 & 41.40 & 42.06 \\
SELECT & \textbf{42.22} & \textbf{48.01} & \textbf{48.94} \\
\bottomrule
\end{tabular}

\end{table}

\paragraph{\rev{Backbone comparison.}}
\rev{Table~\ref{tab:semanticposs_minknet_24k} compares SELECT and baselines with MinkNet at epoch~60 under a 24k-point budget. SELECT achieves 30.37\% mIoU with 21,578 labels.}

\begin{table}[t]
\caption{\textbf{SemanticPOSS with MinkNet at epoch~60.}
The 24,000-point annotation budget includes 6,000 initial labels.}
\label{tab:semanticposs_minknet_24k}
\centering
\begingroup
\small
\setlength{\tabcolsep}{3pt}
\renewcommand{\arraystretch}{1.10}
\begin{tabular*}{\linewidth}{@{\extracolsep{\fill}}lrr@{}}
\toprule
Method & Point labels & mIoU (\%) \\
\midrule
Random          & 24,000 & 25.06 \\
Entropy         & 24,000 & 20.77 \\
Margin~\cite{joshi2009multiclass} & 18,474 & 25.15 \\
VCD (Annotator) & 24,000 & 21.34 \\
BADGE           & 24,000 & 23.83 \\
Core-Set        & 24,000 & 21.08 \\
\midrule
\textbf{SELECT}  & 21,578 & \textbf{30.37} \\
\bottomrule
\end{tabular*}
\endgroup

\end{table}

\paragraph{\rev{Results on nuScenes.}}
\rev{At epoch~60 with 24,000 labels, SELECT reaches 29.65\% mIoU, compared with 25.34\% for Random and 26.27\% for Margin. Initial SPVCNN voxelization differs across these runs (supplementary Sec.~\ref{sec:supp_baselines}).}

\begin{figure*}[t]
    \centering
    
\includegraphics[width=0.85\textwidth]{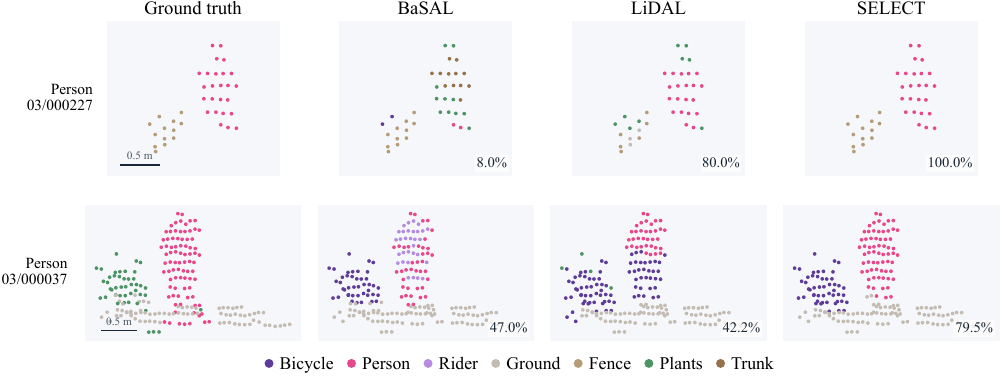}
    \caption{\textbf{Qualitative comparison on SemanticPOSS.} These examples were selected retrospectively to illustrate favorable SELECT predictions. Columns show ground truth and predictions from BaSAL, LiDAL, and SELECT with SPVCNN at epoch~60. Colors indicate semantic classes; percentages are target-class IoU within each crop. All runs use a 24,000-point annotation budget, but actual label counts and training protocols differ.}
    \label{fig:qualitative_semanticposs}
\end{figure*}

\paragraph{\rev{Qualitative results.}}
\rev{Figure~\ref{fig:qualitative_semanticposs} shows two retrospectively selected SemanticPOSS crops at epoch~60. SELECT achieves person IoUs of 100.0\% and 79.5\%, compared with 80.0\% and 42.2\% for LiDAL and 8.0\% and 47.0\% for BaSAL.}

\paragraph{\rev{Computational cost.}}
\rev{The 24k-budget SELECT run takes 18.73 minutes for acquisition, including inference, candidate construction, screening, and joint selection. In separate runs, LESS teacher and student training took about 35.5 hours, while SELECT took about 5.8 hours overall. These measurements cover different training scopes; further timings and memory use are in supplementary Sec.~\ref{sec:supp_result_details}.}

\section{Ablation Study}
\label{sec:ablation}
\paragraph{\rev{Effect of joint selection.}}
\rev{Table~\ref{tab:ablation} compares the joint objective with component variants and fixed-order sequential selection. Under a 24k-point budget, SELECT achieves 48.94\% mIoU with 21,278 labels, compared with 44.96\% using 24,000 labels for sequential selection. Actual label counts also differ across component variants.}

\begin{table}[t]
\caption{Comparison of SELECT, variants without $R_I$, $D$, or $G$, variants using only $R_I$, $D$, or $G$, and sequential selection on SemanticPOSS with SPVCNN. Time is total query time in minutes.}
\label{tab:ablation}
\centering
\small
\setlength{\tabcolsep}{2pt}
\begin{tabular}{lrrrr}
\toprule
Variant & Points & mIoU & Rare4 & Time \\
\midrule
SELECT & 21,278 & 48.94 & 35.58 & 18.73 \\
Without $R_I$ & 24,000 & 39.25 & 21.37 & 11.06 \\
Without $D$ & 19,484 & 46.21 & 32.06 & 18.34 \\
Without $G$ & 20,255 & 47.75 & 32.88 & 16.21 \\
\midrule
$R_I$ only & 18,440 & 44.21 & 25.28 & 129.61 \\
$D$ only & 18,453 & 42.22 & 26.19 & 11.02 \\
$G$ only & 21,291 & 43.59 & 29.02 & 12.59 \\
Sequential & 24,000 & 44.96 & 26.59 & 96.86 \\
\bottomrule
\end{tabular}

\end{table}

\paragraph{\rev{Effect of intensity weighting.}}
\rev{SELECT achieves 2.26 percentage points higher mIoU and 3.30 points higher Rare4 IoU than the variant with unit intensity weights (supplementary Table~\ref{tab:class_results}). Differences are computed before rounding. The runs use 21,278 and 19,448 labels, respectively, in different environments.}

\paragraph{Effect of the screening rule.}
Table~\ref{tab:resources} compares screening rules with Stage~2 intensity weights fixed to one.
Submodular screening takes 4.19 minutes over five rounds, versus over seven hours for the tested $k$-means variants. GMM has only one recorded round.

\begin{table}[!htbp]
\caption{\textbf{Stage 1 screening comparison.} Stage~2 uses unit intensity weights. Timing sums the screening phase and excludes inference and feature preparation.}
\label{tab:resources}
\centering
\small
\setlength{\tabcolsep}{3.5pt}
\begin{tabular}{lrrr}
\toprule
Screening rule & Rounds & Stage 1 time & mIoU \\
\midrule
Uniform & 5 & \textbf{6.87\,s} & 45.21 \\
$k$-means & 5 & 7.51\,h & \textbf{46.04} \\
$k$-means++ & 5 & 9.52\,h & 44.36 \\
GMM & 1 & 32.00\,h & \missing \\
\textbf{Stochastic submodular} & 5 & 4.19\,min & 45.92 \\
\bottomrule
\end{tabular}

\end{table}

\paragraph{\rev{Local prediction and intensity cues.}}
\rev{Prediction disagreement can occur within a single ground-truth class (Fig.~\ref{fig:stage2_prediction_disagreement}), while higher intensity variation is associated with more points near detected depth jumps after reweighting (Fig.~\ref{fig:stage2_intensity_depth}). These observational diagnostics motivate the two cues; details are in supplementary Sec.~\ref{sec:supp_result_details}.}

\section{Conclusion}
\label{sec:conclusion}

\rev{SELECT performs two-stage active LiDAR semantic segmentation under a point-label budget. Representative screening reduces the candidate pool using backbone features. Joint selection combines intensity-weighted coverage, local predicted-class diversity, and acquired label counts by marginal gain per new label. Training uses unlabeled points as spatial context. The frozen objective is monotone submodular. SELECT reaches 48.94\% mIoU with 21,278 labels on SemanticPOSS. Additional seeds and comparisons at matched actual label counts are needed to assess robustness and isolate component contributions.}

\paragraph{Limitations.}
\rev{Predicted class counts may be biased; intensity variation depends on acquisition conditions and does not directly measure semantic error. Scan quotas can leave budget unused, and point counts omit annotation time. Further evaluation should measure annotation time, rare-class IoU, and computational cost under explicit protocols.}

\clearpage
\section*{LLM Usage Statement}
ChatGPT (OpenAI) was used to assist with writing and editing, reviewing related work and citations, coding and implementation, experiments, and data preparation.
The research ideas, method, figures and other visual materials, and all claims and conclusions are the authors' own.
The authors take full responsibility for the content of the paper.

{\small
\bibliographystyle{ieeenat_fullname}
\bibliography{references}

\begin{thebibliography}{27}
\providecommand{\natexlab}[1]{#1}
\providecommand{\url}[1]{\texttt{#1}}
\expandafter\ifx\csname urlstyle\endcsname\relax
  \providecommand{\doi}[1]{doi: #1}\else
  \providecommand{\doi}{doi: \begingroup \urlstyle{rm}\Url}\fi

\bibitem[Ash et~al.(2020)Ash, Zhang, Krishnamurthy, Langford, and
  Agarwal]{ash2020badge}
Jordan~T. Ash, Chicheng Zhang, Akshay Krishnamurthy, John Langford, and Alekh
  Agarwal.
\newblock Deep batch active learning by diverse, uncertain gradient lower
  bounds.
\newblock In \emph{Eighth International Conference on Learning Representations
  ({ICLR} 2020)}. OpenReview.net, 2020.

\bibitem[Behley et~al.(2019)Behley, Garbade, Milioto, Quenzel, Behnke,
  Stachniss, and Gall]{behley2019semantickitti}
Jens Behley, Martin Garbade, Andres Milioto, Jan Quenzel, Sven Behnke, Cyrill
  Stachniss, and J{\"{u}}rgen Gall.
\newblock {SemanticKITTI}: {A} dataset for semantic scene understanding of
  {LiDAR} sequences.
\newblock In \emph{2019 {IEEE/CVF} International Conference on Computer Vision,
  {ICCV} 2019, Seoul, Korea (South), October 27 - November 2, 2019}, pages
  9296--9306. {IEEE}, 2019.

\bibitem[Choy et~al.(2019)Choy, Gwak, and Savarese]{choy20194d}
Christopher~B. Choy, JunYoung Gwak, and Silvio Savarese.
\newblock {4D} spatio-temporal {ConvNets}: Minkowski convolutional neural
  networks.
\newblock In \emph{{IEEE} Conference on Computer Vision and Pattern
  Recognition, {CVPR} 2019, Long Beach, CA, USA, June 16-20, 2019}, pages
  3075--3084. Computer Vision Foundation / {IEEE}, 2019.

\bibitem[Fong et~al.(2022)Fong, Mohan, Hurtado, Zhou, Caesar, Beijbom, and
  Valada]{fong2022panoptic}
Whye~Kit Fong, Rohit Mohan, Juana~Valeria Hurtado, Lubing Zhou, Holger Caesar,
  Oscar Beijbom, and Abhinav Valada.
\newblock Panoptic {nuScenes}: {A} large-scale benchmark for {LiDAR} panoptic
  segmentation and tracking.
\newblock \emph{{IEEE} Robotics Autom. Lett.}, 7\penalty0 (2):\penalty0
  3795--3802, 2022.

\bibitem[Gal et~al.(2017)Gal, Islam, and Ghahramani]{gal2017deep}
Yarin Gal, Riashat Islam, and Zoubin Ghahramani.
\newblock Deep {Bayesian} active learning with image data.
\newblock In \emph{Proceedings of the 34th International Conference on Machine
  Learning, {ICML} 2017, Sydney, NSW, Australia, 6-11 August 2017}, pages
  1183--1192. {PMLR}, 2017.

\bibitem[Hu et~al.(2022{\natexlab{a}})Hu, Yang, Fang, Guo, Leonardis, Trigoni,
  and Markham]{hu2022sqn}
Qingyong Hu, Bo Yang, Guangchi Fang, Yulan Guo, Ales Leonardis, Niki Trigoni,
  and Andrew Markham.
\newblock {SQN:} weakly-supervised semantic segmentation of large-scale {3D}
  point clouds.
\newblock In \emph{Computer Vision - {ECCV} 2022 - 17th European Conference,
  Tel Aviv, Israel, October 23-27, 2022, Proceedings, Part {XXVII}}, pages
  600--619. Springer, 2022{\natexlab{a}}.

\bibitem[Hu et~al.(2022{\natexlab{b}})Hu, Bai, Zhang, Wang, Sun, Fu, and
  Tai]{hu2022lidal}
Zeyu Hu, Xuyang Bai, Runze Zhang, Xin Wang, Guangyuan Sun, Hongbo Fu, and
  Chiew{-}Lan Tai.
\newblock {LiDAL}: Inter-frame uncertainty based active learning for {3D}
  {LiDAR} semantic segmentation.
\newblock In \emph{Computer Vision - {ECCV} 2022 - 17th European Conference,
  Tel Aviv, Israel, October 23-27, 2022, Proceedings, Part {XXVII}}, pages
  248--265. Springer, 2022{\natexlab{b}}.

\bibitem[Iyer and Bilmes(2019)]{iyer2019memoization}
Rishabh~K. Iyer and Jeffrey~A. Bilmes.
\newblock A memoization framework for scaling submodular optimization to large
  scale problems.
\newblock In \emph{The 22nd International Conference on Artificial Intelligence
  and Statistics, {AISTATS} 2019, 16-18 April 2019, Naha, Okinawa, Japan},
  pages 2340--2349. {PMLR}, 2019.

\bibitem[Joshi et~al.(2009)Joshi, Porikli, and
  Papanikolopoulos]{joshi2009multiclass}
Ajay~J. Joshi, Fatih Porikli, and Nikolaos Papanikolopoulos.
\newblock Multi-class active learning for image classification.
\newblock In \emph{2009 {IEEE} Computer Society Conference on Computer Vision
  and Pattern Recognition {(CVPR} 2009), 20-25 June 2009, Miami, Florida,
  {USA}}, pages 2372--2379. {IEEE} Computer Society, 2009.

\bibitem[Kong et~al.(2023)Kong, Ren, Pan, and Liu]{kong2023lasermix}
Lingdong Kong, Jiawei Ren, Liang Pan, and Ziwei Liu.
\newblock {LaserMix} for semi-supervised {LiDAR} semantic segmentation.
\newblock In \emph{{IEEE/CVF} Conference on Computer Vision and Pattern
  Recognition, {CVPR} 2023, Vancouver, BC, Canada, June 17-24, 2023}, pages
  21706--21716. {IEEE}, 2023.

\bibitem[Liu et~al.(2022)Liu, Zhou, Qi, Gong, Su, and Anguelov]{liu2022less}
Minghua Liu, Yin Zhou, Charles~R. Qi, Boqing Gong, Hao Su, and Dragomir
  Anguelov.
\newblock {LESS:} label-efficient semantic segmentation for {LiDAR} point
  clouds.
\newblock In \emph{Computer Vision - {ECCV} 2022 - 17th European Conference,
  Tel Aviv, Israel, October 23-27, 2022, Proceedings, Part {XXXIX}}, pages
  70--89. Springer, 2022.

\bibitem[Mao et~al.(2024{\natexlab{a}})Mao, Maharana, Iyer, and
  Guo]{mao2024stone}
Ruiyu Mao, Sarthak~Kumar Maharana, Rishabh~K. Iyer, and Yunhui Guo.
\newblock {STONE:} {A} submodular optimization framework for active {3D} object
  detection.
\newblock In \emph{Advances in Neural Information Processing Systems 37: Annual
  Conference on Neural Information Processing Systems 2024, NeurIPS 2024,
  Vancouver, BC, Canada, December 10 - 15, 2024}, 2024{\natexlab{a}}.

\bibitem[Mao et~al.(2024{\natexlab{b}})Mao, Xu, and Guo]{mao2024neat}
Ruiyu Mao, Ouyang Xu, and Yunhui Guo.
\newblock Inconsistency-based data-centric active open-set annotation.
\newblock In \emph{Thirty-Eighth {AAAI} Conference on Artificial Intelligence,
  {AAAI} 2024, Thirty-Sixth Conference on Innovative Applications of Artificial
  Intelligence, {IAAI} 2024, Fourteenth Symposium on Educational Advances in
  Artificial Intelligence, {EAAI} 2024, February 20-27, 2024, Vancouver,
  Canada}, pages 4180--4188. {AAAI} Press, 2024{\natexlab{b}}.

\bibitem[Mao et~al.(2026)Mao, Zhang, Ruozzi, and Guo]{mao2026lh3d}
Ruiyu Mao, Baoming Zhang, Nicholas Ruozzi, and Yunhui Guo.
\newblock Learnability-driven submodular optimization for active roadside {3D}
  detection.
\newblock In \emph{{IEEE/CVF} Conference on Computer Vision and Pattern
  Recognition, {CVPR} 2026, Denver, CO, USA, June 3-7, 2026}, pages
  11579--11588. Computer Vision Foundation, 2026.

\bibitem[Mirzasoleiman et~al.(2015)Mirzasoleiman, Badanidiyuru, Karbasi,
  Vondr{\'{a}}k, and Krause]{mirzasoleiman2015lazier}
Baharan Mirzasoleiman, Ashwinkumar Badanidiyuru, Amin Karbasi, Jan
  Vondr{\'{a}}k, and Andreas Krause.
\newblock Lazier than lazy greedy.
\newblock In \emph{Proceedings of the Twenty-Ninth {AAAI} Conference on
  Artificial Intelligence, January 25-30, 2015, Austin, Texas, {USA}}, pages
  1812--1818. {AAAI} Press, 2015.

\bibitem[Pan et~al.(2020)Pan, Gao, Mei, Geng, Li, and
  Zhao]{pan2020semanticposs}
Yancheng Pan, Biao Gao, Jilin Mei, Sibo Geng, Chengkun Li, and Huijing Zhao.
\newblock {SemanticPOSS}: {A} point cloud dataset with large quantity of
  dynamic instances.
\newblock In \emph{{IEEE} Intelligent Vehicles Symposium, {IV} 2020, Las Vegas,
  NV, USA, October 19 - November 13, 2020}, pages 687--693. {IEEE}, 2020.

\bibitem[Samet et~al.(2024)Samet, Puy, Sim{\'{e}}oni, and
  Marlet]{samet2024milan}
Nermin Samet, Gilles Puy, Oriane Sim{\'{e}}oni, and Renaud Marlet.
\newblock {MILAN:} milli-annotations for {Lidar} semantic segmentation.
\newblock \emph{CoRR}, abs/2407.15797, 2024.

\bibitem[Sener and Savarese(2018)]{sener2018coreset}
Ozan Sener and Silvio Savarese.
\newblock Active learning for convolutional neural networks: {A} core-set
  approach.
\newblock In \emph{6th International Conference on Learning Representations,
  {ICLR} 2018, Vancouver, BC, Canada, April 30 - May 3, 2018, Conference Track
  Proceedings}. OpenReview.net, 2018.

\bibitem[Shao et~al.(2022)Shao, Luo, Liu, Chen, Yang, Lu, and
  Xiao]{shao2022ssdr}
Feifei Shao, Yawei Luo, Ping Liu, Jie Chen, Yi Yang, Yulei Lu, and Jun Xiao.
\newblock Active learning for point cloud semantic segmentation via
  spatial-structural diversity reasoning.
\newblock In \emph{{MM} '22: The 30th {ACM} International Conference on
  Multimedia, Lisboa, Portugal, October 10 - 14, 2022}, pages 2575--2585.
  {ACM}, 2022.

\bibitem[Tang et~al.(2020)Tang, Liu, Zhao, Lin, Lin, Wang, and
  Han]{tang2020searching}
Haotian Tang, Zhijian Liu, Shengyu Zhao, Yujun Lin, Ji Lin, Hanrui Wang, and
  Song Han.
\newblock Searching efficient {3D} architectures with sparse point-voxel
  convolution.
\newblock In \emph{Computer Vision - {ECCV} 2020 - 16th European Conference,
  Glasgow, UK, August 23-28, 2020, Proceedings, Part {XXVIII}}, pages 685--702.
  Springer, 2020.

\bibitem[Tran et~al.(2019)Tran, Do, Reid, and Carneiro]{tran2019bgadl}
Toan Tran, Thanh{-}Toan Do, Ian~D. Reid, and Gustavo Carneiro.
\newblock Bayesian generative active deep learning.
\newblock In \emph{Proceedings of the 36th International Conference on Machine
  Learning, {ICML} 2019, 9-15 June 2019, Long Beach, California, {USA}}, pages
  6295--6304. {PMLR}, 2019.

\bibitem[Wei et~al.(2024)Wei, Lin, and Caesar]{wei2023basal}
Jiarong Wei, Yancong Lin, and Holger Caesar.
\newblock {BaSAL}: Size-balanced warm start active learning for {LiDAR}
  semantic segmentation.
\newblock In \emph{{IEEE} International Conference on Robotics and Automation,
  {ICRA} 2024, Yokohama, Japan, May 13-17, 2024}, pages 18258--18264. {IEEE},
  2024.

\bibitem[Wei et~al.(2015)Wei, Iyer, and Bilmes]{wei2015submodularity}
Kai Wei, Rishabh~K. Iyer, and Jeff~A. Bilmes.
\newblock Submodularity in data subset selection and active learning.
\newblock In \emph{Proceedings of the 32nd International Conference on Machine
  Learning, {ICML} 2015, Lille, France, 6-11 July 2015}, pages 1954--1963.
  JMLR.org, 2015.

\bibitem[Wu et~al.(2021)Wu, Liu, Huang, Lee, Su, Huang, and Hsu]{wu2021redal}
Tsung{-}Han Wu, Yueh{-}Cheng Liu, Yu{-}Kai Huang, Hsin{-}Ying Lee, Hung{-}Ting
  Su, Ping{-}Chia Huang, and Winston~H. Hsu.
\newblock {ReDAL}: Region-based and diversity-aware active learning for point
  cloud semantic segmentation.
\newblock In \emph{2021 {IEEE/CVF} International Conference on Computer Vision,
  {ICCV} 2021, Montreal, QC, Canada, October 10-17, 2021}, pages 15490--15499.
  {IEEE}, 2021.

\bibitem[Xie et~al.(2023)Xie, Li, Guo, Liu, and Cheng]{xie2023annotator}
Binhui Xie, Shuang Li, Qingju Guo, Chi~Harold Liu, and Xinjing Cheng.
\newblock Annotator: {A} generic active learning baseline for {LiDAR} semantic
  segmentation.
\newblock In \emph{Advances in Neural Information Processing Systems 36: Annual
  Conference on Neural Information Processing Systems 2023, NeurIPS 2023, New
  Orleans, LA, USA, December 10 - 16, 2023}, 2023.

\bibitem[Xu et~al.(2023)Xu, Yuan, Zhao, Zhang, and Gao]{xu2023hpal}
Zongyi Xu, Bo Yuan, Shanshan Zhao, Qianni Zhang, and Xinbo Gao.
\newblock Hierarchical point-based active learning for semi-supervised point
  cloud semantic segmentation.
\newblock In \emph{{IEEE/CVF} International Conference on Computer Vision,
  {ICCV} 2023, Paris, France, October 1-6, 2023}, pages 18052--18062. {IEEE},
  2023.

\bibitem[Ye et~al.(2023)Ye, Yin, Fu, Lin, and Pan]{ye2023multigranularity}
Shanding Ye, Zhe Yin, Yongjian Fu, Hu Lin, and Zhijie Pan.
\newblock A multi-granularity semisupervised active learning for point cloud
  semantic segmentation.
\newblock \emph{Neural Comput. Appl.}, 35\penalty0 (21):\penalty0 15629--15645,
  2023.

\end{thebibliography}
}
\clearpage
\maketitlesupplementary
This supplement provides implementation and theoretical details, annotation protocols, and additional experimental results.

\section*{Organization of the Supplementary Material}

\noindent\textbf{\ref{sec:select_method_details}. Additional Method Details}\par
\vspace{2pt}
\noindent\begin{tabularx}{\linewidth}{@{}l@{\hspace{0.6em}}X@{}}
\ref{sec:select_objective_scales} & Objective Scales \\
\ref{sec:select_implementation} & Implementation Settings \\
\ref{sec:select_procedure} & Acquisition Procedure \\
\ref{sec:select_objective_properties} & Objective Properties \\
\ref{sec:select_computational_cost} & Computational Cost \\
\end{tabularx}
\par\vspace{6pt}

\noindent\textbf{\ref{sec:supp_experiment_setup}. Additional Experimental Setup}\par
\vspace{2pt}
\noindent\begin{tabularx}{\linewidth}{@{}l@{\hspace{0.6em}}X@{}}
\ref{sec:supp_baselines} & Active Learning and Data Selection Baselines \\
\end{tabularx}
\par\vspace{6pt}

\noindent\textbf{\ref{sec:supplementary_results}. Additional Experimental Results}\par
\vspace{2pt}
\noindent\begin{tabularx}{\linewidth}{@{}l@{\hspace{0.6em}}X@{}}
\ref{sec:supp_result_details} & Additional Acquisition and Diagnostic Details \\
\end{tabularx}
\par\vspace{6pt}

\noindent\textbf{\ref{sec:cross_dataset_results}. Cross-Dataset Results and Component Comparisons}\par
\vspace{2pt}
\noindent\begin{tabularx}{\linewidth}{@{}l@{\hspace{0.6em}}X@{}}
\ref{sec:kitti_spv} & SemanticKITTI with SPVCNN \\
\ref{sec:nusc_spv} & nuScenes with SPVCNN \\
\ref{sec:components_budget} & Effect of Joint Selection and Objective Components \\
\end{tabularx}
\par\vspace{6pt}

\noindent\textbf{\ref{sec:additional_completed}. Additional Backbone and History Comparisons}\par
\vspace{2pt}
\noindent\begin{tabularx}{\linewidth}{@{}l@{\hspace{0.6em}}X@{}}
\ref{sec:kitti_mink_completed} & SemanticKITTI with MinkNet \\
\ref{sec:nusc_completed} & nuScenes with MinkNet \\
\ref{sec:without_history_completed} & Effect of Acquired-Class History \\
\end{tabularx}
\par\vspace{6pt}

\noindent\textbf{\ref{sec:supp_visualizations}. Additional Visualizations}\par
\vspace{2pt}
\noindent\begin{tabularx}{\linewidth}{@{}l@{\hspace{0.6em}}X@{}}
\ref{sec:supp_training_trajectories} & Training trajectories \\
\ref{sec:supp_query_locations_cost} & Query locations and annotation cost \\
\ref{sec:supp_local_predictions} & Local predictions and a failure case \\
\end{tabularx}
\par\vspace{6pt}

\appendix
\section{Additional Method Details}
\label{sec:select_method_details}

This section gives the objective scales, implementation settings, selection procedure, and computational cost.
It uses the notation of the Background and Method sections.

\subsection{Objective Scales}
\label{sec:select_objective_scales}

For a round with $b_t>0$ and a nonempty screened pool, we use
\begin{equation}
\begin{aligned}
    Z_R &=
    d\log\left(1+\frac{b_t}{d\tau_f}\right),\\
    Z_D &= b_t,\\
    Z_G &=
    \max\left\{
        \sum_{c=1}^{C}
        \log\left(1+\frac{b_t/C}{\tau_b+h_{t,c}}\right),
        10^{-12}
    \right\}.
\end{aligned}
    \label{eq:select_objective_scales}
\end{equation}
These constants set the relative scales of the three components.
They use the initial allowance $b_t$ for the round and remain fixed as the remaining budget decreases.
In particular, $Z_G$ is a scaling convention and need not upper-bound $G$ for every history vector.
A round with zero allowance or no eligible candidates returns an empty query before computing these quantities.

\subsection{Implementation Settings}
\label{sec:select_implementation}

\paragraph{Query construction.}
For SemanticPOSS, we use a nominal query-cell step of $0.25$\,m and a network resolution of $0.05$\,m.
Query coordinates are quantized by truncation toward zero.
For a scan with $J_i$ occupied cells, let
$o_{i,(1)}\geq\cdots\geq o_{i,(J_i)}$
be the raw cell occupancies in descending order.
The retained-member width is
\begin{equation}
    H_i=\min\{100,o_{i,(k_i)}\},
    \qquad k_i=\lfloor0.01J_i\rfloor.
    \label{eq:select_retention_vidth}
\end{equation}
The evaluated scans satisfy $k_i\geq1$.
A cell retains at most $H_i$ members in the order produced by the grouping operation.
Thus, $\mathcal{T}_v$ can be a strict subset of the points in a physical cell.
Point identity is preserved, so retained members that already belong to $L_t$ remain available for feature averaging but are excluded from $\mathcal{U}_{t,v}$.

\paragraph{Feature preparation and parameters.}
We use $d_p=16$, giving $d=32$ feature coordinates, with $K=512$, $\varepsilon=0.1$, $\tau_f=16$, and $\tau_b=100$.
Let $\mathbf{e}_{ij}\in\mathbb{R}^{d_f}$ be the feature of point $\mathbf{p}_{ij}$.
We average these features over each candidate's retained members:
\begin{equation}
    \mathbf{z}_v=
    \frac{1}{|\mathcal{T}_v|}
    \sum_{(i,j)\in\mathcal{T}_v}\mathbf{e}_{ij}.
    \label{eq:method_mean_feature}
\end{equation}
A fixed Gaussian projection $\mathbf{M}\in\mathbb{R}^{d_f\times d_p}$ reduces the feature dimension.
We separate positive and negative coordinates and normalize the result:
\begin{equation}
\begin{aligned}
    \mathbf{g}_v
        &= [(\mathbf{z}_v\mathbf{M})_+,
            (-\mathbf{z}_v\mathbf{M})_+],\\
    \boldsymbol{\phi}_v
        &= \frac{\mathbf{g}_v}
        {\max(\|\mathbf{g}_v\|_1,\eta)},
\end{aligned}
    \label{eq:method_projected_feature}
\end{equation}
where $(\cdot)_+$ is the element-wise positive part, $\eta>0$ prevents division by zero, and $d=2d_p$ is the resulting dimension.
The projection is linear, so we apply it to point features before gathering and averaging retained members.
This avoids constructing a retained-member feature tensor in the original dimension $d_f$.
\paragraph{Stochastic screening.}
Let $\mathcal{V}_{t,i}$ be the eligible candidates in scan $i$ and $A_{t,i}$ the screened set.
Given the current set $A$, we sample uniformly without replacement from $\mathcal{V}_{t,i}\setminus A$.
For $\varepsilon\in(0,1)$, the sample size is
\begin{equation}
    \min\left\{
        |\mathcal{V}_{t,i}\setminus A|,
        \left\lceil
        \frac{|\mathcal{V}_{t,i}|}{K}
        \log\frac{1}{\varepsilon}
        \right\rceil
    \right\}.
    \label{eq:method_sample_size}
\end{equation}
A swap-delete array maintains remaining candidates during stochastic screening.

\paragraph{Local statistics and history.}
Feature averages use $\mathcal{T}_v$.
Annotation costs, intensity dispersion, predicted class counts, and local diversity use $\mathcal{U}_{t,v}$.
Intensity is taken directly from raw point records and divided by $255$ for SemanticPOSS.
An appropriate raw-intensity scale is required when using another dataset.
The intensity-weight denominator uses the maximum dispersion over the entire screened pool $\mathcal{C}_t$ and stays fixed during selection.
Ignored labels or predictions do not contribute to class counts.
Feature-coverage sums and selected-batch class counts reset each round. Acquired true-label counts accumulate across rounds.

\paragraph{Selection details.}
A lazy priority queue stores previously evaluated marginal gains divided by the fixed costs $\kappa_v$.
After accepting a candidate, the feature and predicted-class sums are updated, and other candidates from its scan become infeasible.
A candidate that exceeds the remaining budget is skipped without truncating its membership.
Selection continues while a feasible candidate remains, including when its marginal gain is zero.
Exact density ties are resolved by the majority-class logit score in ascending order and then by candidate identity.
The logit score is used only for tie breaking and has no additive weight in $F_t$.
The main method uses one joint-selection branch and trains with acquired human labels.

\subsection{Acquisition Procedure}
\label{sec:select_procedure}
Algorithm~\ref{alg:select} summarizes one frozen SELECT acquisition round.

\begin{algorithm}[t]
\caption{One SELECT acquisition round}
\label{alg:select}
\begin{algorithmic}[1]
\Require Model $f_\theta$, labeled and unlabeled indices $L_t,U_t$, history $\mathbf{h}_t$, capped allowance $b_t$
\If{$b_t=0$}
    \State \Return $Q_t=\varnothing$
\EndIf
\State Freeze model and initialize $\mathcal{C}_t\gets\varnothing$
\For{each training scan $\mathcal{P}_i$}
    \State Extract features $\mathbf{e}_{ij}$ and predictions $\hat c_{ij}$
    \State Construct $\mathcal{T}_v$, $\mathcal{U}_{t,v}$, and $\kappa_v$
    \State Form $\mathcal{V}_{t,i}$ from candidates with $\kappa_v>0$
    \State Compute $\boldsymbol{\phi}_v$ over retained members
    \State Screen $A_{t,i}$ using $R_0$ and stochastic greedy
    \State $\mathcal{C}_t\gets\mathcal{C}_t\cup A_{t,i}$
\EndFor
\If{$\mathcal{C}_t=\varnothing$}
    \State \Return $Q_t=\varnothing$
\EndIf
\State Compute and freeze $\omega_v,n_{v,c},D_v,Z_R,Z_D,Z_G$
\State Set $S\gets\varnothing$ and initialize the lazy priority queue
\While{a candidate is feasible under the cost and scan constraints}
    \State Find a feasible $v$ with largest current $\rho_t(v\mid S)$
    \State $S\gets S\cup\{v\}$
    \State Update feature/class sums, remaining budget, and scan availability
\EndWhile
\State $Q_t\gets\bigcup_{v\in S}\mathcal{U}_{t,v}$
\State Request true labels for $Q_t$
\State Update $L_{t+1}$ and $\mathbf{h}_{t+1}$
\State \Return $Q_t$ and updated supervision
\end{algorithmic}
\end{algorithm}

\subsection{Objective Properties}
\label{sec:select_objective_properties}

For nonnegative fixed coefficients $s_v$ and a positive offset $\tau$, consider
$g(S)=\log(1+\sum_{v\in S}s_v/\tau)$.
For $u\notin S$, its marginal gain is
\begin{equation}
    \Delta_g(u\mid S)=
    \log\left(1+\frac{s_u}{\tau+\sum_{v\in S}s_v}\right).
    \label{eq:select_log_marginal}
\end{equation}
The gain is nonnegative and decreases as $S$ grows.
Thus, $g$ is normalized, nonnegative, monotone, and submodular.

Applying this argument coordinate-wise gives the properties of $R_0$ with $s_v=\phi_{vk}$ and $R_I$ with $s_v=\omega_v\phi_{vk}$.
It also applies to each class term of $G$ with $s_v=n_{v,c}$ and offset $\tau_b+h_{t,c}$.
Each voxel score $D_v$ in Eq.~\eqref{eq:method_prediction_diversity} is fixed and nonnegative, so $D$ is modular.
Nonnegative component weights and fixed positive scales preserve these properties, so $F_t$ is normalized, nonnegative, monotone, and submodular.
The argument assumes $\tau_f,\tau_b>0$ and $C\geq2$.

For the per-scan screening problem with $|\mathcal{V}_{t,i}|>K$, stochastic greedy gives
\begin{equation}
    \mathbb{E}[R_0(A_{t,i})]
    \geq
    (1-1/e-\varepsilon)
    \max_{\substack{A\subseteq\mathcal{V}_{t,i}\\|A|\leq K}}
    R_0(A)
    \label{eq:select_screening_guarantee}
\end{equation}
under the stated uniform sampling rule~\cite{mirzasoleiman2015lazier}.
When $|\mathcal{V}_{t,i}|\leq K$, retaining all candidates maximizes the monotone screening objective.
This guarantee concerns the fixed feature-screening objective.
It does not provide a segmentation-accuracy bound or a guarantee for the final density-based selection under point costs and per-scan constraints.

\subsection{Computational Cost}
\label{sec:select_computational_cost}

Let $M_{t,i}=|\mathcal{V}_{t,i}|$.
For $M_{t,i}>K$, screening evaluates at most
\begin{equation}
    K\left\lceil
        \frac{M_{t,i}}{K}\log\frac{1}{\varepsilon}
    \right\rceil
    \label{eq:select_screening_evaluations}
\end{equation}
candidate marginals.
Each evaluation uses $d$ coordinates, giving
$O(d[M_{t,i}\log(1/\varepsilon)+K])$
objective-evaluation work.
The screening representation and active indices require $O(M_{t,i}d+M_{t,i})$ storage per scan.
These bounds exclude model inference, feature preparation, and retained raw-point records.

For joint selection, let $M_t=|\mathcal{C}_t|$ and $m_t=|S_t|$.
A conservative bound is $O(M_t(m_t+1))$ marginal evaluations, including queue initialization.
Each evaluation uses $O(d+C)$ work, and queue operations add logarithmic overhead.
The stored candidate features and class counts use $O(M_t(d+C))$ space, with additional storage for the heap and raw-point identities.
Lazy evaluation can reduce repeated evaluations in practice.
The cost bounds do not make the full acquisition pipeline linear, and measured acquisition time includes inference, candidate preparation, screening, and joint selection.
\section{Additional Experimental Setup}
\label{sec:supp_experiment_setup}

\subsection{Active Learning and Data Selection Baselines}
\label{sec:supp_baselines}

\paragraph{Active learning and data selection.}
The \emph{voxel-based baselines}---random sampling, entropy sampling, margin sampling~\cite{joshi2009multiclass}, and \rev{voxel confusion degree (VCD)}---share a voxel-query interface following Annotator~\cite{xie2023annotator}; VCD adapts its selection score.
On SemanticPOSS, we also evaluate budget-aligned adaptations grouped by their original query units: \emph{region-based methods} ReDAL~\cite{wu2021redal}, SSDR-AL~\cite{shao2022ssdr}, LiDAL~\cite{hu2022lidal}, and BaSAL~\cite{wei2023basal}; \emph{point-based} HPAL~\cite{xu2023hpal}; and MILAN~\cite{samet2024milan}, which combines \emph{scan selection} with sparse annotation of cluster representatives.
BaSAL uses its size-balanced initialization.
LiDAL, MILAN, and HPAL use method-specific pseudo-labels or label propagation in their adapted training protocols.

\paragraph{\rev{SQN.}}
\rev{Our SQN adaptation uses a sparse Minkowski encoder with a four-scale semantic query head. The same 6,000-point initialization is supplemented with 18,000 uniformly sampled points without replacement; all 24,000 labels are available from epoch~1, with no active acquisition.}

\paragraph{\rev{nuScenes voxelization.}}
\rev{SELECT uses integer-exact initial SPVCNN voxel coordinates under mixed precision. Initial voxelization and training settings differ across runs, so the nuScenes comparison evaluates the complete configurations.}

\paragraph{Implementation details.}
We evaluate SELECT with SPVCNN~\cite{tang2020searching} and MinkNet~\cite{choy20194d}; the reported backbone runs use different acquisition configurations.
Both backbones use point coordinates and LiDAR return intensity.
The network voxel size and query grid spacing are 0.05\,m and 0.25\,m, respectively.
For SPVCNN, Stage~1 retains at most 512 candidates per scan using stochastic greedy screening with $\varepsilon=0.1$; Stage~2 selects by marginal gain per new point label.
We train for 60 epochs with batch size two, SGD, momentum 0.9, weight decay $10^{-4}$, and learning-rate warmup followed by cosine decay.
SELECT runs use seed 1234.
The 24k SemanticPOSS SELECT run with SPVCNN uses automatic mixed precision on one NVIDIA RTX A5000 GPU.
Comparison tables report single-run scores evaluated at epoch~60.

\paragraph{Active learning setup.}
SELECT and the four shared voxel baselines start with the same 6,000 labeled raw points within each dataset.
Acquisition rounds are scheduled at epochs 4, 7, 10, 13, and 16, with at most 6,000 new point labels per round, limited by the remaining total budget.
\rev{Total annotation budgets are 12,000, 18,000, and 24,000 on SemanticPOSS and 24,000 on nuScenes and SemanticKITTI.}
The 24,000-point annotation budget is approximately $0.0143\%$ of SemanticPOSS raw training points.
SELECT queries at most one whole candidate per scan per round, requesting labels for all its retained unlabeled points.
Each acquired human-labeled raw point counts once, including initialization and acquired ignore-label points; we report actual counts alongside the budgets.
\rev{Tables~\ref{tab:actual_counts_classwise} and~\ref{tab:actual_counts_budgets} list actual label counts for the main comparison and the budget sweep.}

\paragraph{\rev{Evaluation.}}
\rev{We map voxel predictions back to raw points and compute mIoU over 13 SemanticPOSS, 16 nuScenes, and 19 SemanticKITTI valid classes. Points with ignored ground-truth labels are excluded. Ignore predictions on valid points count as false negatives for the corresponding ground-truth classes. IoUs are averaged over the fixed valid class set; a class with zero union contributes zero.}

\section{Additional Experimental Results}
\label{sec:supplementary_results}

SELECT achieves the highest reported mIoU at each SemanticPOSS budget in the main-paper sweep (Table~\ref{tab:budget_comparison}): 42.22\%, 48.01\%, and 48.94\% at 12k, 18k, and 24k, respectively.
At the 24k budget, SELECT uses 21,278 labels, or approximately 0.0127\% of the raw training points.
Table~\ref{tab:actual_counts_budgets} reports actual counts; the compared runs use different label counts and training protocols.

\subsection{Additional Acquisition and Diagnostic Details}
\label{sec:supp_result_details}

\paragraph{Acquisition cost.}
On SemanticPOSS, acquisition takes 7.68, 14.70, and 18.73 minutes at the 12k, 18k, and 24k budgets, respectively, including model inference, candidate construction, screening, and joint selection.
The 24k run takes 5.80 hours end to end, with a sampled GPU memory peak of 10,826\,MiB for the training process tree.

\paragraph{Query voxel size.}
The $0.25$\,m query setting achieves the highest mIoU among the four tested sizes in Table~\ref{tab:query_voxel_size}.
It uses 21,278 labels, compared with 24,000 for the $0.50$\,m and $0.75$\,m settings, at the same $0.05$\,m network resolution. \begin{table}[t]
\caption{\textbf{Query voxel size on SemanticPOSS.} Single-seed SPVCNN runs use a 24k annotation budget, including initialization, and a fixed $0.05$\,m network voxel size. Points are the actual labels acquired.} \label{tab:query_voxel_size}
\centering \small \begin{tabular}{lrr} \toprule Query size (m) & Points & mIoU (\%) \\ \midrule 0.05 & 18,567 & 47.05 \\ 0.25 & 21,278 & \textbf{48.94} \\ 0.50 & 24,000 & 47.01 \\ 0.75 & 24,000 & 43.54 \\ \bottomrule \end{tabular}  \end{table} \paragraph{Class-support ablation.}
On SemanticPOSS with SPVCNN, SELECT and the variant without $G$ achieve Rare4 IoU of $35.58\%$ and $32.88\%$, and overall mIoU of $48.94\%$ and $47.75\%$, respectively, at epoch~60 (Table~\ref{tab:ablation}).
Both runs use seed 1234 and a 24k annotation budget, but acquire 21,278 and 20,255 labels, respectively.

\paragraph{Local prediction and intensity cues.}
We analyze query voxels with at least two retained unlabeled points on 20 SemanticPOSS validation scans.
After reweighting by unlabeled point count and range, mean within-voxel prediction error is 60.88\% for high-disagreement voxels and 14.46\% for voxels with zero predicted-class diversity.
For the intensity comparison, we define a depth-discontinuity proxy using LiDAR rays within $0.25^\circ$ whose return distances differ by at least 1\,m.
After reweighting the low-variation group by unlabeled point count, range, and dominant ground-truth class, the mean within-voxel share of points near these flagged discontinuities is 29.18\% for high-variation voxels and 11.96\% for low-variation voxels.
These reweighted comparisons are observational, not causal.

\paragraph{Intensity-weighted coverage.}
Table~\ref{tab:class_results} reports 48.94\% mIoU and 35.58\% Rare4 IoU for SELECT, 2.26 and 3.30 percentage points above the unit-weight reference using unrounded scores.
Mean Rare4 recall increases from 38.68\% to 42.93\%.
The largest listed class gain is on garbage can, from 0.05\% to 9.46\% IoU; person and rider IoUs are slightly lower.
The two runs acquire different numbers of labels and use different environments.

\begin{table}[t]
\caption{\textbf{Selected class-wise results on SemanticPOSS.} Values are IoU except for Rare4 recall, the mean per-class recall. Rare4 averages rider, pole, garbage can, and cone/stone; Rare5 additionally includes traffic sign. The unit-weight variant sets $\omega_v=1$. \rev{Differences are computed before rounding and reported in percentage points.} SELECT and the unit-weight variant use 21,278 and 19,448 labels, respectively, and run in different environments.}
\label{tab:class_results}
\centering
\small
\setlength{\tabcolsep}{5pt}
\begin{tabular}{lrrr}
\toprule
Class & Unit weights & SELECT & Difference \\
\midrule
Person & 55.34 & 54.45 & $-0.89$ \\
Rider & 54.61 & 54.34 & $-0.27$ \\
Pole & 33.03 & 34.82 & \rev{$+1.80$} \\
Traffic-sign & 32.88 & 35.68 & $+2.80$ \\
Garbage-can & 0.05 & 9.46 & $+9.41$ \\
Cone/stone & 41.45 & 43.70 & $+2.25$ \\
\midrule
mIoU & 46.69 & 48.94 & \rev{$+2.26$} \\
Rare4 & 32.28 & 35.58 & $+3.30$ \\
Rare5 & 32.40 & 35.60 & $+3.20$ \\
Rare4 recall & 38.68 & 42.93 & $+4.25$ \\
\bottomrule
\end{tabular}

\end{table}

\begin{table}[t]
\caption{\rev{\textbf{Actual label counts for the SemanticPOSS class-wise comparison.} Counts include initialization and acquired ignore labels. All methods use a 24k-point budget.}}
\label{tab:actual_counts_classwise}
\centering
\revcolor
\small
\begin{tabular}{lrr}
\toprule
Method & Actual labels & mIoU (\%) \\
\midrule
Random & 24,000 & 38.25 \\
Entropy & 24,000 & 35.95 \\
BADGE & 24,000 & 41.72 \\
Core-Set & 24,000 & 38.16 \\
ReDAL & 24,000 & 43.38 \\
SQN & 24,000 & 43.83 \\
BaSAL & 23,997 & 42.06 \\
LiDAL & 24,000 & 44.33 \\
MILAN-derived & 6,109 & 30.88 \\
SSDR-AL & 23,992 & 33.28 \\
Margin & 18,468 & 41.18 \\
VCD (Annotator) & 23,998 & 36.73 \\
HPAL & 24,000 & 34.67 \\
LaserMix & 24,000 & 20.69 \\
SELECT & 21,278 & 48.94 \\
\bottomrule
\end{tabular}

\end{table}

\begin{table}[t]
\caption{\rev{\textbf{Actual label counts for the SemanticPOSS budget sweep.} Counts correspond to the single-seed epoch-60 scores in Table~\ref{tab:budget_comparison} and include initialization and acquired ignore labels.}}
\label{tab:actual_counts_budgets}
\centering
\revcolor
\small
\begin{tabular}{lrrr}
\toprule
Method & 12k budget & 18k budget & 24k budget \\
\midrule
Random & 12,000 & 18,000 & 24,000 \\
ReDAL & 12,000 & 18,000 & 24,000 \\
LiDAL & 12,000 & 18,000 & 24,000 \\
BaSAL & 11,998 & 17,998 & 23,997 \\
SELECT & 12,000 & 18,000 & 21,278 \\
\bottomrule
\end{tabular}

\end{table}

\section{Cross-Dataset Results and Component Comparisons}
\label{sec:cross_dataset_results}

Under a 24k-point annotation budget, SELECT achieves the highest reported SPVCNN mIoU on SemanticPOSS, SemanticKITTI, and nuScenes: 48.94\%, 27.21\%, and 29.65\%, respectively (main-paper Table~\ref{tab:budget_results} and Tables~\ref{tab:kitti_spv}--\ref{tab:nusc_spv}).
It uses 21,278 labels on SemanticPOSS and 24,000 on each of the other two datasets.
All results in this section are single-run raw-point evaluations at epoch~60; budgets include initialization.

\subsection{SemanticKITTI with SPVCNN}
\label{sec:kitti_spv}

SELECT ranks first among the six methods in Table~\ref{tab:kitti_spv} at 27.21\% mIoU, above Random (19.07\%) and slightly above Margin (27.09\%).
All six runs acquire 24,000 labels; the close SELECT--Margin result is a single-run comparison.

\begin{table}[ht]
\caption{\textbf{SemanticKITTI with SPVCNN at epoch~60.} Single-run, 19-class results under a 24k-point budget. Labels denote actual human point labels, including initialization. BADGE and Core-Set are point-label adaptations. Bold marks the highest mIoU.}
\label{tab:kitti_spv}
\centering
\small
\begin{tabular}{lrr}
\toprule
Method & Labels & mIoU (\%) \\
\midrule
Random & 24,000 & 19.07 \\
Entropy & 24,000 & 18.87 \\
Margin & 24,000 & 27.09 \\
BADGE adaptation & 24,000 & 21.77 \\
Core-Set adaptation & 24,000 & 16.96 \\
\midrule
SELECT & 24,000 & \textbf{27.21} \\
\bottomrule
\end{tabular}
\end{table}

\subsection{nuScenes with SPVCNN}
\label{sec:nusc_spv}

SELECT exceeds all eight SPVCNN baselines in Table~\ref{tab:nusc_spv}.
Its 29.65\% mIoU is 4.31 percentage points above Random and 3.38 points above the strongest listed baseline, Margin.
The runs use different initial voxelization and method-specific training settings (Sec.~\ref{sec:supp_baselines}).

\begin{table}[ht]
\caption{\textbf{nuScenes with SPVCNN at epoch~60.} Single-run, 16-class results under a 24k-point budget. Initial voxelization and training adaptations differ across runs. Labels denote actual human point labels. Bold marks the highest mIoU.}
\label{tab:nusc_spv}
\centering
\small
\begin{tabular}{lrr}
\toprule
Method & Labels & mIoU (\%) \\
\midrule
Random & 24,000 & 25.34 \\
Entropy & 24,000 & 25.41 \\
Margin & 24,000 & 26.27 \\
VCD (Annotator) & 24,000 & 25.56 \\
BADGE adaptation & 24,000 & 23.94 \\
Core-Set adaptation & 24,000 & 21.34 \\
LiDAL & 24,000 & 25.65 \\
SSDR-AL & 23,977 & 21.67 \\
\midrule
SELECT & 24,000 & \textbf{29.65} \\
\bottomrule
\end{tabular}
\end{table}

\subsection{Effect of Joint Selection and Objective Components}
\label{sec:components_budget}

On SemanticPOSS with SPVCNN, joint SELECT attains 48.94\% mIoU and 35.58\% Rare4 IoU, compared with 44.96\% and 26.59\% for sequential selection (Table~\ref{tab:components_budget}).
The gains are 3.98 and 8.99 percentage points, respectively, with fewer labels (21,278 versus 24,000).
Sequential selection splits each round's budget equally among $R_I$, $D$, and $G$ in that order, without reallocating unused shares.
SELECT also has the highest reported mIoU and Rare4 among the listed configurations.
The runs share a 24k-point budget but have different actual label counts; the unit-weight reference in Table~\ref{tab:class_results} also uses a different environment.

\begin{table}[ht]
\caption{\textbf{SemanticPOSS component comparisons under a 24k-point budget.} Single-run SPVCNN results at epoch~60. Actual label counts differ. The unit-weight variant sets $\omega_v=1$ and uses the reference run in Table~\ref{tab:class_results}. Rare4 averages rider, pole, garbage can, and cone/stone IoUs.}
\label{tab:components_budget}
\centering
\small
\begin{tabular}{lrrr}
\toprule
Variant & Labels & mIoU & Rare4 \\
\midrule
SELECT & 21,278 & \textbf{48.94} & \textbf{35.58} \\
Unit intensity weights & 19,448 & 46.69 & 32.28 \\
Without $R_I$ & 24,000 & 39.25 & 21.37 \\
Without $D$ & 19,484 & 46.21 & 32.06 \\
Without $G$ & 20,255 & 47.75 & 32.88 \\
Sequential & 24,000 & 44.96 & 26.59 \\
\bottomrule
\end{tabular}
\end{table}

\section{Additional Backbone and History Comparisons}
\label{sec:additional_completed}

We extend the SPVCNN comparisons with MinkNet and examine the role of acquired-class history.
All results are evaluated at epoch~60 under a 24k-point annotation budget, including initialization.

\subsection{SemanticKITTI with MinkNet}
\label{sec:kitti_mink_completed}

SELECT reaches 16.47\% mIoU, the second-highest score among the eight methods in Table~\ref{tab:kitti_mink_completed}.
It improves over Random (12.58\%) and all five other non-Margin baselines.
Margin achieves the highest mIoU at 21.68\%.
Rankings vary across the reported configurations, which also differ in their acquisition settings.

\begin{table}[htbp]
\caption{\textbf{SemanticKITTI with MinkNet at epoch~60.} Single-run, 19-class results under a 24k-point budget. VCD, BADGE, Core-Set, and SSDR-AL use adapted protocols. Actual label counts differ. Bold marks the highest mIoU.}
\label{tab:kitti_mink_completed}
\centering
\small
\begin{tabular}{lrr}
\toprule
Method & Labels & mIoU (\%) \\
\midrule
Random & 24,000 & 12.58 \\
Entropy & 24,000 & 11.45 \\
Margin & 24,000 & \textbf{21.68} \\
VCD (Annotator) & 23,999 & 13.35 \\
BADGE adaptation & 24,000 & 12.78 \\
Core-Set adaptation & 24,000 & 10.93 \\
SSDR-AL & 23,933 & 11.09 \\
SELECT & 24,000 & 16.47 \\
\bottomrule
\end{tabular}
\end{table}

\subsection{nuScenes with MinkNet}
\label{sec:nusc_completed}

On nuScenes, SELECT reaches 17.18\% mIoU with MinkNet, above BaSAL (11.82\%) and SSDR-AL (15.18\%).
MILAN-derived and Random attain 19.10\% and 17.78\%, respectively (Table~\ref{tab:nusc_mink_completed}).
The favorable SPVCNN ranking in Table~\ref{tab:nusc_spv} therefore does not extend uniformly to MinkNet.
These comparisons use method-specific adaptations, label counts, and supervision protocols.

\begin{table}[htbp]
\caption{\textbf{nuScenes with MinkNet at epoch~60.} Single-run, 16-class results under a 24k-point budget. Actual label counts and method-specific supervision protocols differ. Bold marks the highest mIoU.}
\label{tab:nusc_mink_completed}
\centering
\small
\begin{tabular}{lrr}
\toprule
Method & Labels & mIoU (\%) \\
\midrule
Random & 24,000 & 17.78 \\
BaSAL & 24,000 & 11.82 \\
SSDR-AL & 23,988 & 15.18 \\
MILAN-derived & 24,000 & \textbf{19.10} \\
SELECT & 24,000 & 17.18 \\
\bottomrule
\end{tabular}
\end{table}

\subsection{Effect of Acquired-Class History}
\label{sec:without_history_completed}

SELECT reaches 48.94\% mIoU and 35.58\% Rare4 IoU, compared with 44.39\% and 27.85\% when the history input to $G$ is set to zero (Table~\ref{tab:without_history_completed}).
Computed before rounding, the gains are 4.56 and 7.73 percentage points, respectively, while SELECT acquires fewer labels (21,278 versus 22,014).
This ablation retains all three objective terms and sets the acquired-history vector to zero only when evaluating $G$; $Z_G$ is still computed from the actual acquired history.
It therefore differs from removing $G$ entirely.
These are single-run SPVCNN comparisons with different actual label counts.

\begin{table}[htbp]
\caption{\textbf{Effect of acquired-class history on SemanticPOSS.} Single-run SPVCNN results at epoch~60 under a 24k-point budget. All three terms are retained; only the history used within $G$ is zeroed in the ablation. Actual labels differ.}
\label{tab:without_history_completed}
\centering
\small
\begin{tabular}{lrrr}
\toprule
Variant & Labels & mIoU & Rare4 \\
\midrule
SELECT & 21,278 & \textbf{48.94} & \textbf{35.58} \\
Without acquired history & 22,014 & 44.39 & 27.85 \\
\bottomrule
\end{tabular}
\end{table}

\section{Additional Visualizations}
\label{sec:supp_visualizations}
These figures show training trajectories, query locations, annotation costs, and local predictions on SemanticPOSS. SELECT uses intensity-weighted coverage, prediction diversity, and class support.

\subsection{Training trajectories}
\label{sec:supp_training_trajectories}
\begin{figure*}[t]
\centering
\includegraphics[width=\textwidth]{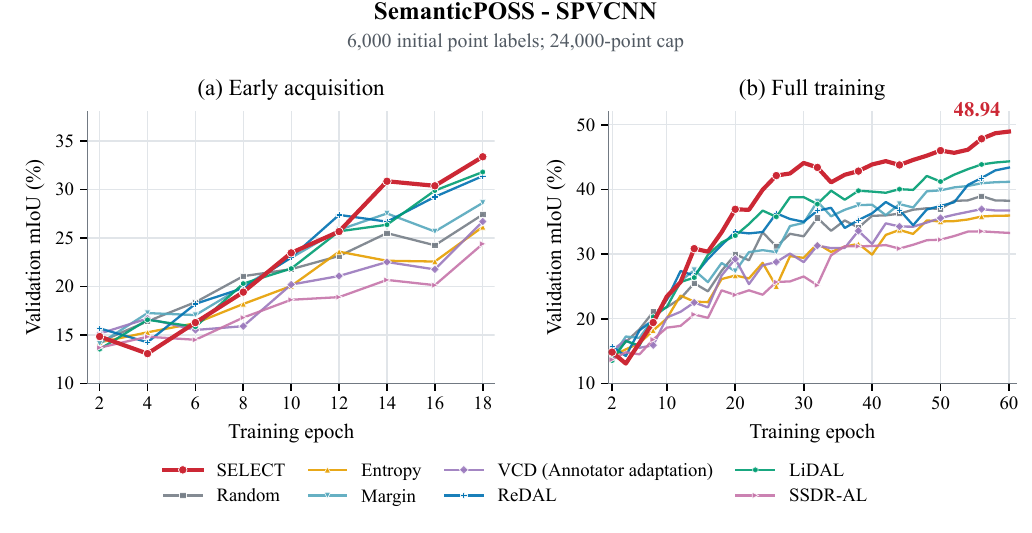}
\caption{\textbf{Learning trajectories on SemanticPOSS.} SPVCNN runs with seed~1234, 6,000 initial labels, and a 24k total annotation cap. Left: epochs~2--18. Right: training through epoch~60. SELECT uses 21,278 point labels in total, including initialization. Lines connect validation evaluations. Acquisition rounds are scheduled at epochs~4/7/10/13/16, and training continues afterward. Adapted baseline protocols and actual label counts differ (Table~\ref{tab:actual_counts_classwise}). Methods are trained independently.}
\label{fig:supp_learning}
\end{figure*}

Figure~\ref{fig:supp_learning} shows validation performance during acquisition and subsequent training: SELECT reaches 33.37\% mIoU at epoch~18 and 48.94\% at epoch~60, the highest final score among the eight plotted methods. The methods are trained independently and have different validation scores before the first acquisition round.

\subsection{Query locations and annotation cost}
\label{sec:supp_query_locations_cost}
\begin{figure*}[t]
\centering
\includegraphics[width=0.62\textwidth]{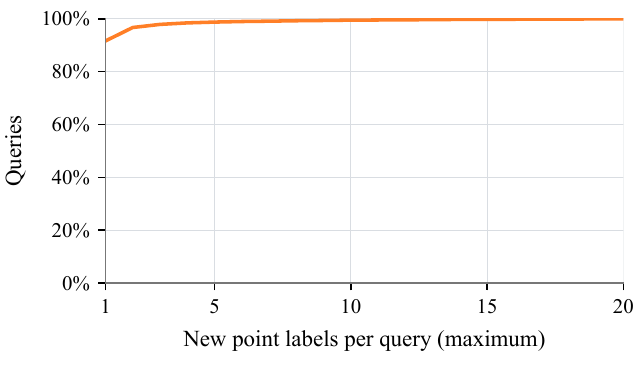}
\caption{\textbf{New labels per SELECT query.} Cumulative distribution over all 12,440 nonempty queries across five rounds. The horizontal axis counts newly labeled raw points per query; initialization is excluded. A total of 91.61\% of queries add one label. The plot shows costs up to 20 new labels; five queries have costs of 21--23. These costs are not physical voxel occupancy counts or measurements of human annotation time.}
\label{fig:supp_query_cost}
\end{figure*}

\begin{figure*}[t]
\centering
\includegraphics[width=\textwidth]{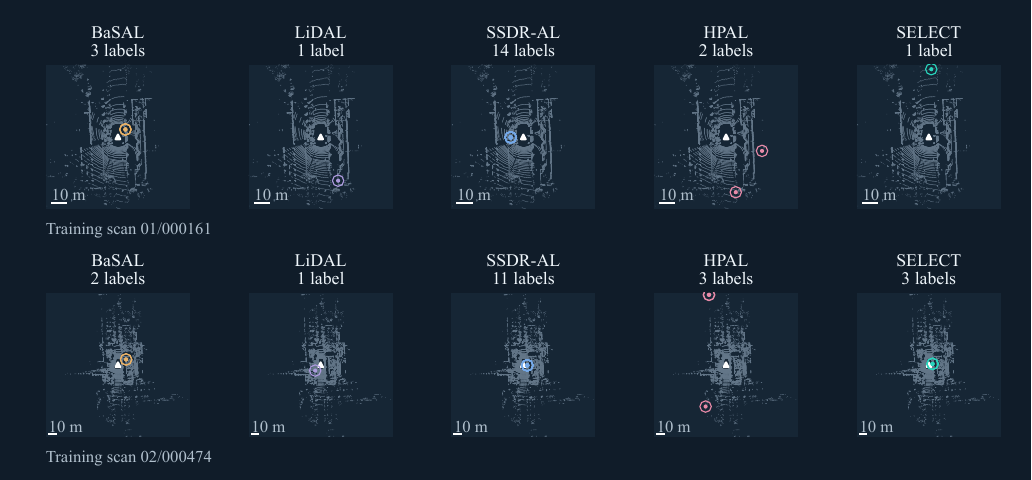}
\caption{\textbf{Actual first-round query locations.} SemanticPOSS training scans 01/000161 (top) and 02/000474 (bottom), epoch~4. Each row uses the same top-down view. Colored markers show newly acquired raw-point locations and are enlarged for visibility; the white triangle marks the sensor. Scan IDs were selected at fixed positions in the sorted set of scans queried by all five methods, without filtering by class or performance. Counts above panels refer only to the displayed scan and round. They reflect each complete acquisition policy and its evolving model.}
\label{fig:supp_query_locations}
\end{figure*}

Across five rounds, 12,440 queries add 15,278 labels; 91.61\% add one label (Fig.~\ref{fig:supp_query_cost}). Costs count only retained unlabeled members, excluding initialization. A physical cell may also contain already labeled or unretained points. Figure~\ref{fig:supp_query_locations} shows actual first-round queries in two training scans, illustrating spatially selective supervision with full-scan input.

\subsection{Local predictions and a failure case}
\label{sec:supp_local_predictions}
\begin{figure*}[t]
\centering
\includegraphics[width=\textwidth]{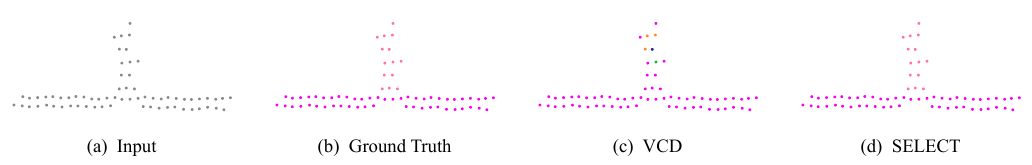}\\[0.5em]
\includegraphics[width=\textwidth]{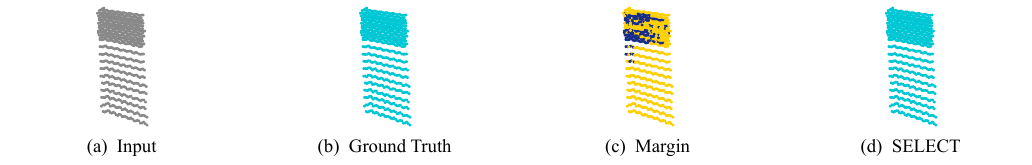}
\caption{\textbf{Small-object prediction examples.} Top: cone/stone at 20.3\,m in validation scan 03/000009; local target-class IoU is 0.0\% for VCD and 100.0\% for SELECT. Bottom: traffic sign at 6.6\,m in scan 03/000103; local IoU is 0.0\% for Margin and 100.0\% for SELECT. Columns show input, ground truth, the named baseline, and SELECT. Actual labels are 23,998 (VCD), 18,468 (Margin), and 21,278 (SELECT), including initialization. Scores refer to these fixed crops, not class IoU over the full validation set. These selected examples are not comparisons at identical actual label counts.}
\label{fig:supp_small_objects}
\end{figure*}

\begin{figure*}[t]
\centering
\includegraphics[width=\textwidth]{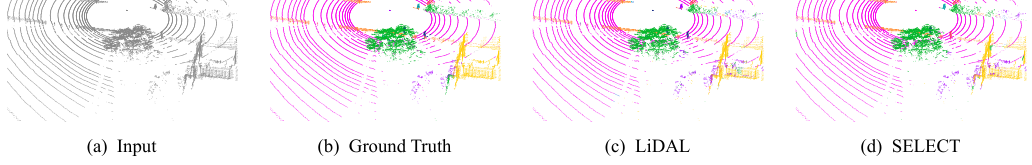}\\[0.5em]
\includegraphics[width=\textwidth]{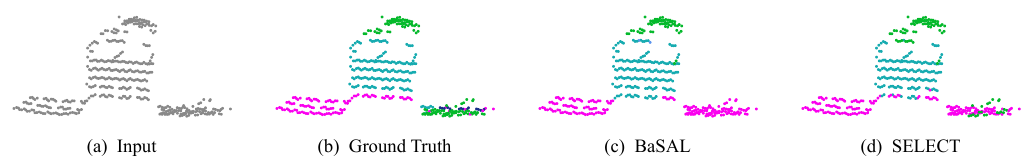}
\caption{\textbf{Garbage-can prediction with scene context.} Validation scan 03/000272, target range 8.1\,m. Top: scene view comparing LiDAL and SELECT. Bottom: fixed local crop comparing BaSAL and SELECT. On this crop, target-class IoU is 23.0\% for LiDAL, 83.9\% for BaSAL, and 84.2\% for SELECT. Actual labels are 24,000 (LiDAL), 23,997 (BaSAL), and 21,278 (SELECT). The baseline changes between rows as indicated in the panel labels. SELECT and BaSAL obtain similar IoU on this crop.}
\label{fig:supp_garbage_can}
\end{figure*}

\begin{figure*}[t]
\centering
\includegraphics[width=\textwidth]{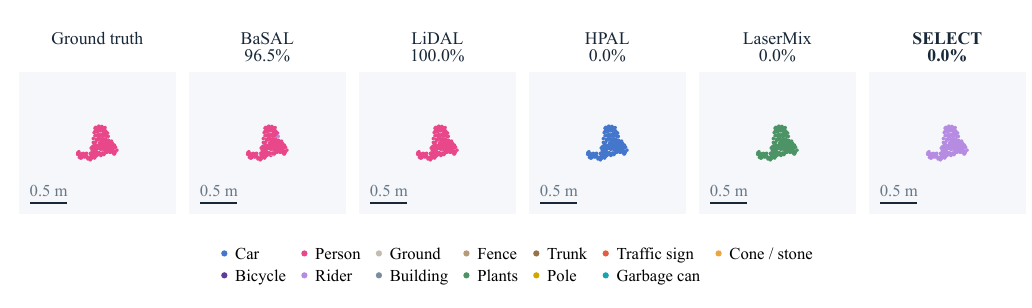}
\caption{\textbf{An illustrative failure case.} Person at 3.3\,m in validation scan 03/000008, with 231 ground-truth target points. Percentages are target-class IoU within the fixed displayed crop: BaSAL 96.5\%, LiDAL 100.0\%, and HPAL, LaserMix, and SELECT 0.0\%. Actual labels are 23,997 (BaSAL), 24,000 (LiDAL/HPAL/LaserMix), and 21,278 (SELECT).}
\label{fig:supp_failure}
\end{figure*}

The selected epoch-60 validation examples are distinct from the main-paper pedestrian cases. SELECT correctly segments the small targets missed by the displayed VCD and Margin predictions (Fig.~\ref{fig:supp_small_objects}). Its garbage-can crop IoU is higher than LiDAL's and close to BaSAL's (Fig.~\ref{fig:supp_garbage_can}). A failure to segment a person (Fig.~\ref{fig:supp_failure}) shows that higher overall mIoU does not guarantee correct predictions for every object. The figures illustrate individual predictions; the tables report aggregate performance.

\end{document}